\documentclass{article}

\PassOptionsToPackage{numbers, compress}{natbib}
\PassOptionsToPackage{nonatbib}{neurips_2020}
\usepackage[nonatbib,preprint]{neurips_2020}
\usepackage[numbers]{natbib}
\usepackage[utf8]{inputenc} % allow utf-8 input
\usepackage[T1]{fontenc}    % use 8-bit T1 fonts
\usepackage{hyperref}       % hyperlinks
\usepackage{url}            % simple URL typesetting
\usepackage{booktabs}       % professional-quality tables
\usepackage{amsfonts}       % blackboard math symbols
\usepackage{nicefrac}       % compact symbols for 1/2, etc.
\usepackage{microtype}      % microtypography
\usepackage{wrapfig}
\usepackage[normalem]{ulem}
\usepackage{lipsum}
\usepackage{blkarray}
\usepackage{todonotes}
\usepackage{comment}
\usepackage{amsmath}
\usepackage{amsfonts}
\usepackage{amssymb}
\usepackage{microtype}
\usepackage{graphicx}
\usepackage{subcaption}
\usepackage{enumitem}
\usepackage{booktabs} % for professional tables
\usepackage{hyperref}
\usepackage{url}
\usepackage{xcolor}
\usepackage{graphicx}
\usepackage{bbm}

\usepackage{cleveref}

\newcommand{\dynamic}[0]{dynamic}
\newcommand{\dynamism}{dynamism}
\newcommand{\states}{\ensuremath{\mathcal{S}}}
\newcommand{\actions}{\ensuremath{\mathcal{A}}}
\newcommand{\transition}{\ensuremath{\mathcal{P}}}

\newcommand{\modifiedtransition}{\ensuremath{\tilde{\mathcal{P}}}}
\newcommand{\policy}{\ensuremath{\pi}}
\newcommand{\optpolicy}{\ensuremath{\pi^*}}
\newcommand{\initialstate}{\ensuremath{\rho}}
\newcommand{\shapedinitialstate}{\ensuremath{\tilde{\rho}}}
\newcommand{\shapedinitialstates}{\ensuremath{\tilde{\rho}_1,...,\tilde{\rho}_k}}

\newcommand{\mdp}{\ensuremath{M}}
\newcommand{\trainmdp}{\ensuremath{\tilde{M}}}
\newcommand{\shapedmdps}{\ensuremath{\tilde{M}_1,...,{\tilde{M}_k}}}
\newcommand{\testmdp}{\ensuremath{M_{\text{test}}}}

\newcommand{\infnorm}[1]{\left\lVert#1\right\rVert_{\infty}}

\title{Ecological Reinforcement Learning}

\author{John D. Co-Reyes$^{*}$ \hspace{5mm} Suvansh Sanjeev\thanks{equal contribution.} \\ \textbf{Glen Berseth} \hspace{5mm}   \textbf{Abhishek Gupta} \hspace{5mm} \textbf{Sergey Levine} \\
University of California Berkeley \\
\texttt{\{jcoreyes,suvansh\}@berkeley.edu}
} 

\begin{document}

\maketitle
\newtheorem{theorem}{Theorem}[section]
\newtheorem{corollary}{Corollary}[theorem]
\newtheorem{lemma}[theorem]{Lemma}
\newtheorem{definition}{Definition}
\newtheorem{proposition}{Proposition}
\newtheorem{assumption}{Assumption}
\begin{abstract}
Much of the current work on reinforcement learning studies episodic settings, where the agent is reset between trials to an initial state distribution, often with well-shaped reward functions. Non-episodic settings, where the agent must learn through continuous interaction with the world without resets, and where the agent receives only delayed and sparse reward signals, is substantially more difficult, but arguably more realistic considering real-world environments do not present the learner with a convenient ``reset mechanism" and easy reward shaping. In this paper, instead of studying algorithmic improvements that can address such non-episodic and sparse reward settings, we instead study the kinds of environment properties that can make learning under such conditions easier. Understanding how properties of the environment impact the performance of reinforcement learning agents can help us to structure our tasks in ways that make learning tractable. We first discuss what we term ``environment shaping" -- modifications to the environment that provide an alternative to reward shaping, and may be easier to implement.
We then discuss an even simpler property that we refer to as ``dynamism," which describes the degree to which the environment changes independent of the agent's actions and can be measured by environment transition entropy. Surprisingly, we find that even this property can substantially alleviate the challenges associated with non-episodic RL in sparse reward settings. We provide an empirical evaluation on a set of new tasks focused on non-episodic learning with sparse rewards. Through this study, we hope to shift the focus of the community towards analyzing how properties of the environment can affect learning and the ultimate type of behavior that is learned via RL.
\end{abstract}
\section{Introduction}
A central goal in current AI research, especially in reinforcement learning (RL), is to develop algorithms that are general, in the sense that the same method can be used to train an effective model for a wide variety of tasks, problems, and domains. In RL, this means designing algorithms that can solve any Markov decision process (MDP). However, natural intelligent agents -- e.g., humans and animals -- exist in the context of a natural environment. The learned behavior of RL agents cannot be understood separately from the environments that they inhabit any more than brains can be understood separately from the bodies they control. There has been comparatively little study in the field of reinforcement learning towards understanding how properties of the environment impact the learning process for complex RL agents. Many of the environments used in modern reinforcement learning research differ in fundamental ways from the real world. First, standard RL benchmarks, such as the arcade learning environment (ALE)~\citep{Bellemare2013TheAL} and Gym~\citep{Brockman2016OpenAIG} are studied with \emph{episodic} RL, while natural environments are continual and lack a ``reset'' mechanism, requiring an agent to learn through continual non-episodic interaction. 
Second, the benchmark environments are typically static, which means only the agent's own actions substantively impact the world. By contrast, natural environments are stochastic and dynamic: an agent that does nothing will still experience many different states, due to the behavior of natural processes and other creatures. Third, benchmark environments typically present the agent with complex but well-shaped reward functions which provide the primary means of guiding the agent to the right solution. In contrast, natural environments might exhibit sparser rewards, but at the same time provide more scaffolding for the learning process, in the form of other cooperative agents, teachers, and curricula. Training in sparse reward non-episodic environments will be important if we want our agents to continually learn useful behavior in realistic scenarios with less supervision.

We use the term \emph{ecological} reinforcement learning to refer to the study of how the behavior of an RL agent is affected by the environment in which it is situated. We specifically aim to study how the non-episodic RL problem is affected by particular properties of the training environments. We first find that modifying the initial state or dynamics of the training environment in a helpful manner (what we define as ``environment shaping''), can allow effective learning in the non-episodic settings, even from sparse, uninformative rewards. This change to the environment induces a sort of curriculum. Next, and perhaps more surprisingly, we observe that simply increasing the entropy of the transition dynamics (what we define as ``environment dynamism'') can also stabilize reset-free learning, and improve learning both in theory and in practice, without any curriculum. We show that environments that have either of these two properties can be far more conducive to non-episodic learning. This observation is significant, because these properties are often lacking in standard benchmark tasks, but arguably are natural and straightforward to provide (and may even already exist) in real-world settings.

The contribution of this work is an analysis of how the properties of MDPs
-- particularly properties that we believe reflect realistic environments -- can make non-episodic reinforcement learning easier. Our results show that environment shaping can ease the difficulty of exploration by reducing the hitting time of a random policy towards high reward regions. We also show that, alternatively, increasing the stochasticity of the environment without explicit shaping or curricula can make it possible to learn without resets even when it is not feasible in the original less stochastic environment. Crucially, our results show that agents trained under these settings actually perform \emph{better} in the original MDPs, which lack dynamic behavior or environment shaping. We hope that our experimental conclusions will encourage future research that studies how the nature of the environment in which the RL agent is situated can facilitate learning and the emergence of complex skills.
\section{Ecological RL Properties} \label{sec:env_properties}
In contrast to most simulated environments that are used for reinforcement learning experiments~\citep{Brockman2016OpenAIG, Bellemare2013TheAL, deepmindlab}, agents learning in natural environments experience a continual stream of experience, without episode boundaries or reset mechanisms. The typical reward function engineering that is often employed in reinforcement learning experiments is also generally unavailable in the real world, where agents must rely on their own low-level perception to understand the world. Rather than studying reinforcement learning algorithms in the general case, our aim is to study how these aspects of the environment impact the performance of reinforcement learning agents and can enable effective training in non-episodic sparse reward settings. We first describe the non-episodic problem setting, and then introduce environment shaping and environment dynamism, which can mitigate the challenges of non-episodic RL.

\begin{figure*} 
    \vspace{-0.07in}
    \centering
    \includegraphics[width=1.0\textwidth]{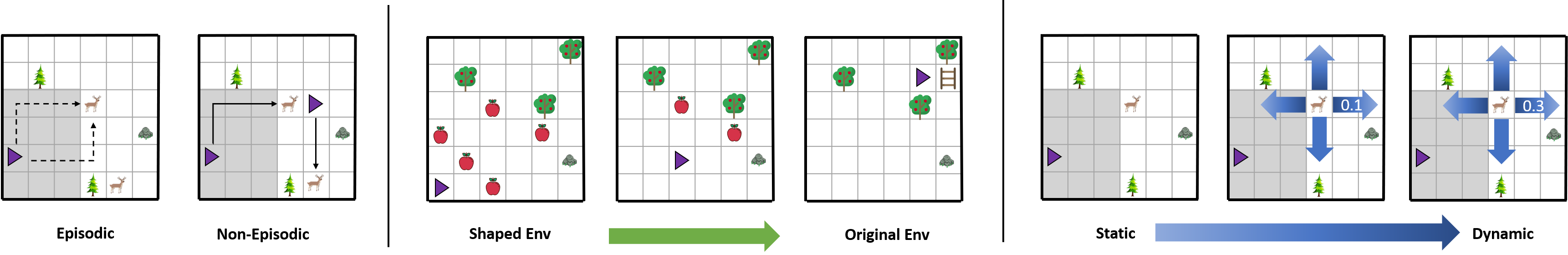}
    \vspace{-0.15in}
    
    \caption{\footnotesize{Left: In an episodic environment, the agent can attempt the task of hunting deer (described in Section \ref{sec:exp_tasks}) multiple times from the same state. In a non-episodic environment, there are no task boundaries and the agent cannot reset to a particular state. The agent instead continually learns across its single lifetime. Middle: Example of environment shaping where fallen apples are initially abundant and easily reachable. Once those are exhausted and the agent has learned apples are valuable it learns to climb the tree to obtain more. Right: Example of varying the \dynamism in the hunting environment where a static one is where the deer are not moving and a more dynamic one is where transitions have higher entropy (the deer move to a random adjacent square with higher probability.}}
    \label{fig:all_properties}
    \vspace{-0.2in}
\end{figure*}

To define the non-episodic setting and these properties (illustrated in Figure \ref{fig:all_properties}) more precisely, we first provide some standard definitions.  The environment is modeled as a Markov decision process (MDP), represented by $M = \lbrace \states, \actions, R, \transition, \gamma, \initialstate \rbrace$, with state space $\states$, action space $\actions$, reward function \hbox{$R: \states \times \actions \rightarrow \mathbb{R}$}, dynamics $\transition: \states \times \actions \times \states \rightarrow [0,1]$, and initial state distribution $\initialstate(s)$. A policy $\pi: \states \times \actions \rightarrow [0,1]$ maps states to an action distribution. The goal is to find the optimal policy $\pi^{*}$ that maximizes the discounted expected return $v^{\pi}(\initialstate) = E_{\pi,s_{0} \sim \initialstate} \sum_{t=0}^T \gamma^{t} R(s_{t}, a_{t})$. We will consider $\trainmdp$ to be MDP that the agent will collect experience from to learn $\pi$.  $\testmdp$ will be the MDP that we evaluate the trained agent on. While we cannot alter $\testmdp$, we will assume that we have some control over the kinds of environments (\trainmdp) we use for training , and we will discuss how manipulating certain properties of these environments can make training substantially easier.

In non-episodic RL, the agent collects a single episode from the training MDP $\trainmdp$, which continues indefinitely ($T=\infty$), without any reset mechanism to bring the agent back to an initial state. Training in this setting will be useful if we want agents that can continually learn in open world environments (such as for the case of a household robot) without having to manually reset the agent and environment back to the exact same initial state. This setting can be 
\begin{wrapfigure}{r}{0.30\textwidth}
  \vspace{-8pt}
  \begin{minipage}[b]{\linewidth}
  \centering
  \includegraphics[width=\linewidth]{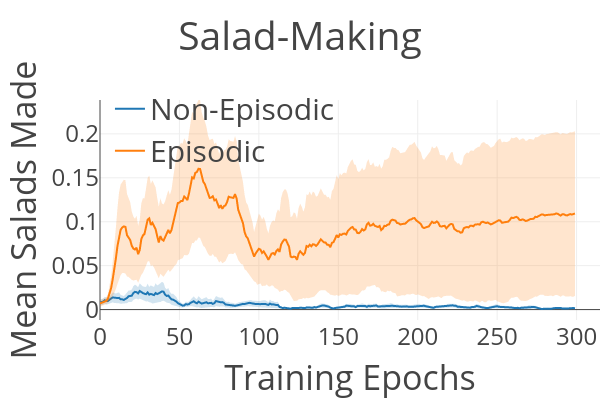}     % performance curve
  \begin{subfigure}{0.45\linewidth}
  \centering\includegraphics[width=0.65\linewidth]{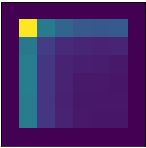}  % heat map episodic
  \caption*{\footnotesize{Non-Episodic}}
  \end{subfigure}%
  \hfill
  \begin{subfigure}{0.45\linewidth}
  \centering\includegraphics[width=0.65\linewidth]{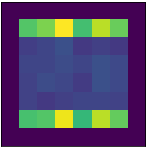}  % heat map non-episodic
  \caption*{\footnotesize{Episodic}}
  \end{subfigure}
  \caption{\footnotesize{Performance curve and state-visitation heat maps for salad-making task (described in Section \ref{sec:exp_tasks}) showing that non-episodic learning is more difficult. The agent struggles to reach any reward and gets stuck in corners.}}
  \label{fig:nonep_vs_ep}
  \end{minipage}
  \vspace{-25pt}
\end{wrapfigure}
difficult because if the environment contains many irreversible states, the agent can become stuck without external help or trapped in local minima. We observe this behavior in our experiments (Figure \ref{fig:nonep_vs_ep}) where we see that the agent trained in the non-episodic version of the task rarely experiences any reward and fails to make any learning progress, becoming trapped in the corners of the state space. 

In practice, a common way to make learning easier, even in the non-episodic setting, is to provide \emph{reward shaping}. However, specifying the task with well shaped rewards in the non-episodic setting may be difficult to provide in the real world. Reward shaping often requires knowledge of privileged state variables (e.g., positions of objects) or the process by which the task must be completed (e.g., required resources). Reward shaping might also introduce bias, since the optimal policy for a shaped reward may not be optimal for the original task reward \citep{Amodei2016ConcretePI}. Instead, it may be more practical to provide a sparse reward signal for the task that is unbiased and easy to measure. The sparse reward non-episodic setting is more challenging but we believe it is more realistic and ultimately more useful if we want to train RL agents which can continually learn in the real world.

\subsection{Environment Shaping} \label{sec:properties_shaping}
Agents in the real world do not learn in a vacuum: humans have a long developmental period and it is reasonable to assume a \emph{cooperative} environment that has been set up to facilitate learning. For humans, this kind of ``scaffolding'' is often provided by other agents (e.g., parents and teachers).  Natural environments might provide automatic scaffolding -- e.g., an animal might first consume readily supplied food but once depleted, must learn to scavenge for its own food (see illustration in Figure~\ref{fig:all_properties}). This kind of \textbf{environment shaping} could serve to guide the learning process, perhaps more easily than reward shaping. We hypothesize that, if we carefully choose the training environment $\trainmdp$, then spare reward non-episodic  RL can be made more tractable.  

Environment shaping corresponds to a choice of training MDP $\trainmdp$ that modifies the initial state distribution to $\shapedinitialstate$ or modifies the dynamics to $\modifiedtransition$. For example, in the hunting task where the goal is to catch deer (Section~\ref{sec:exp_tasks}), $\shapedinitialstate$ would correspond to the presence of easy docile deer in the environment which move towards the agent and make the task of catching deer easier early on in the non-episodic learning process, while in the factory task (Figure \ref{fig:environment_example}), $\modifiedtransition$ would correspond to having cooperative workers that help the agent in the beginning and gradually back off. In either case, shaping is applied during non-episodic training to make learning easier. In the real world we might imagine that parents shape the developmental process of their children by modifying $\shapedinitialstate$ or $\modifiedtransition$. 

To analyze the effect of environment shaping, we build off formalism from \citet{Agarwal2019OptimalityAA, Kakade2002ApproximatelyOA}, which gives the iteration complexity for learning an optimal policy in terms of a mismatch coefficient between the optimal policy state distribution $d^{\optpolicy}_{\initialstate,\transition}(s)$ and the training policy state distribution $d^{\pi}_{\initialstate,\transition}(s)$. If the initial policy state distribution is very different from the optimal policy state distribution, which will be especially true for sparse reward tasks (i.e., if the chance of reaching a reward in a sparse reward task is exponentially low), then the mismatch coefficient can be arbitrarily high, leading to a higher iteration complexity. In altering the initial state distribution or dynamics, environment shaping alters the policy state distribution from $d^{\pi}_{\initialstate,\transition}(s)$  to $d^{\pi}_{\shapedinitialstate,\modifiedtransition}(s)$, pushing it closer to the optimal policy state distribution $d^{\optpolicy}_{\initialstate,\transition}(s)$ and reducing the mismatch coefficient. Intuitively, this means environment shaping can make learning substantially easier by increasing the chance that the agent will visit similar states from $\trainmdp$ with a \emph{bad} initial policy as would be visited from $\mdp$ by a \emph{good} policy. As we discuss in Section~\ref{sec:related_work}, this is closely related to curriculum learning, but applied in the non-episodic setting. In Appendix \ref{appdx:theory_env_shaping} we provide further theoretical analysis on how environment shaping can make non-episodic RL easier.

\textbf{Claim 1}: Environment shaping can enable agents to learn with simple sparse rewards in non-episodic settings, and in some cases result in more proficient policies. Environment shaping can be more effective than reward shaping.

\subsection{Environment Dynamism} \label{sec:propertiesdynamism}
While non-episodic RL can be made easier with environment shaping which is already simpler and more intuitive than reward shaping, surprisingly, similar benefits can be obtained by training the agent in environments where transitions are more stochastic. We call this property \textbf{environment dynamism}. Dynamic environments present their own challenges, but they can also facilitate learning, by automatically exposing the agent to a wide variety of situations. We argue that this dynamism is present in many real-world environments and thus we do not need to purposely train our agents in very controlled and static settings. For example training a trash collecting robot may benefit from a dynamic environment with multiple humans who walk around and drop litter to collect. Dynamics environments can make non-episodic learning easier and may already exist in many scenarios.

We characterize more dynamic training environments $\trainmdp$ as MDPs with transition functions $\modifiedtransition(s'|s,a)$ that have higher entropy for every policy $\pi$ and state $s$, as compared to the original MDP: $\mathcal{H}_{\modifiedtransition,\pi}(\states' |\states=s) \geq \mathcal{H}_{\transition,\pi}(\states' |\states=s) $. In practice, we create dynamic environments by adding dynamic elements that evolve even if the agent does not carry out any coordinated action, such as the deer in Figure \ref{fig:all_properties}, which move randomly with greater probability instead of staying still. Perhaps surprisingly, although on the surface dynamism might appear to make a task harder, it can in fact make non-episodic learning substantially easier. In Appendix \ref{appdx:dynamism_theory}, we demonstrate that more dynamic environments can in effect induce a more uniform `soft' reset distribution in the non-episodic setting. The effect of this will be to increase the state coverage or marginal state entropy of the training policy $\mathcal{H}_{\policy}(\states) $. This will reduce the mismatch coefficient in the worse case. In fact, prior work~\citep{Kakade2002ApproximatelyOA} has shown that a \emph{uniform} reset distribution ensures that the mismatch coefficient is improved in the worse case, since the training policy will have broad coverage if it can reset everywhere. While a fully uniform distribution is difficult to attain in the non-episodic case, the effect of dynamism in broadening state coverage may have a similar effect.

\textbf{Claim 2}: Increasing dynamism alleviates some of the challenges associated with non-episodic learning, by inducing a more uniform state distribution even in the absence of coordinated and intelligent behavior (such as early on in training).

The implications from these hypotheses is that we can facilitate sparse reward non-episodic RL by shaping the environment in a helpful and unbiased manner at the start of the learning process. Even without this shaping, placing our agents in more dynamic environments can also help with the challenges of non-episodic RL.

\section{Experiments} \label{sec:experiments}
To empirically carry out our study of how the properties described in Section~\ref{sec:env_properties} can affect the non-episodic learning problem, we construct different tasks that require non-episodic learning and systematically analyze the impact these properties have on learning dynamics. We considered tasks with sparse ``fundamental drive'' style rewards, which are only provided each time agent has completed the desired task. We train the agent in non-episodic environments, both with and without environment shaping and dynamism, to study the effects of these properties. Crucially, in all cases, we evaluate the agent in the \emph{original} unmodified test MDP $\testmdp$. While a real world study is out of scope for this work, it is evident that many of these properties are present in real world environments. Our code to produce the results and the environments is available at \mbox{\url{https://sites.google.com/view/ecological-rl/home}}

\subsection{Environment and Tasks} \label{sec:exp_tasks}

The five tasks that we evaluate on are depicted in Figure~\ref{fig:environment_example}, with additional details in Appendix \ref{appdx:env_details}. The simulator used for the first four tasks is built on the environment proposed by \citet{gym_minigrid}, which allows us to build compositional tasks that require completing a sequence of subtasks. This setting also allows us to vary environment shaping and dynamism in a controlled way. The last task is from the Unity ML-Agents toolkit \citep{unity_mlagents}, which has a continuous state space. Environments are made non-episodic by removing resets and periodically evaluating the agent on $\testmdp$ during training. Environments are shaped by either placing intermediate resources close to the agent in the beginning, or altering the behavior of other agents and entities to be more cooperative. Environments are made more dynamic by having other agents move randomly with greater probability. While these settings were designed by human designers, in real-world settings we might seek to place our agents in training environments that already have these properties intrinsically. We note that we are not proposing a benchmark set of problems or focusing on particular algorithms, but instead showing through these tasks that environment shaping and environment dynamism can significantly improve the performance of non-episodic RL agents. The tasks are described below.

\begin{figure}[tb] 
\vspace{-8pt}
    \centering
    \includegraphics[width=0.83\columnwidth]{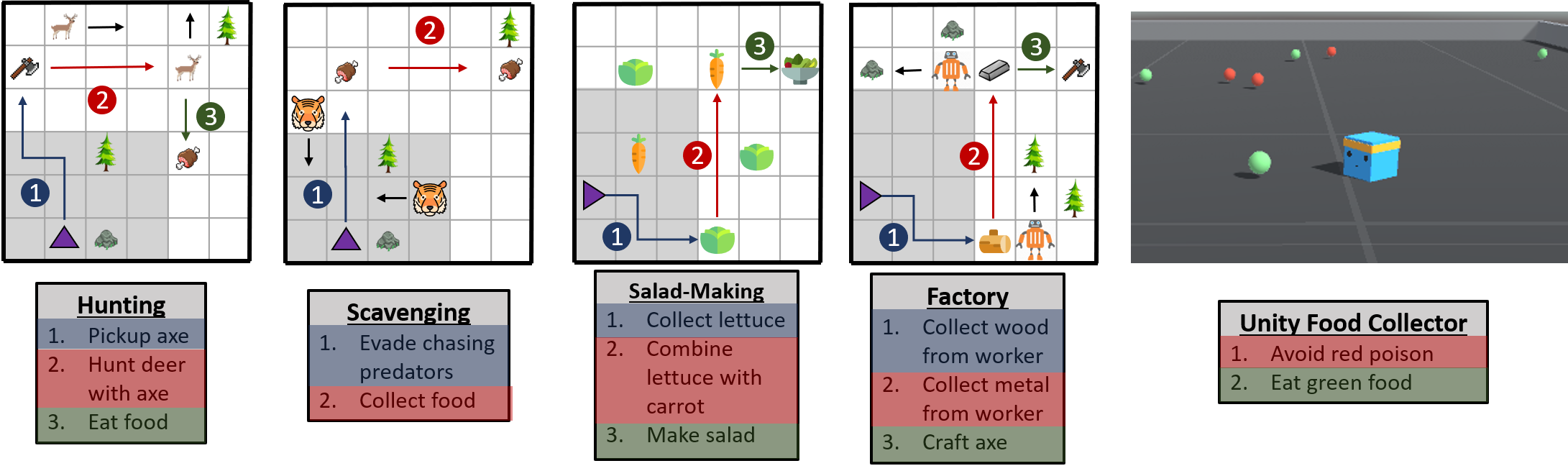}
\vspace{-8pt}
    \caption{\footnotesize{Tasks in our experimental analysis. In the first four tasks, the agent (purple triangle) receives a local observation (shaded gray square) and must interact with objects in the correct sequences to complete the task. In the Unity food collection task (right), the agent must learn to collect green food and avoid red poison.}}
    \label{fig:environment_example}
\vspace{-10pt}
\end{figure} 

\textbf{Hunting:} The task is to pick up an axe and then catch a deer.
\textbf{Dynamism} determines the probability with which deer will move to a random adjacent square. \textbf{Environment shaping} adds easy deer (which move towards the agent) to the initial state of the original environment which only contained hard deer that move away from the agent. As the population of easy deer are eaten, only hard deer are left which creates a natural curriculum. This task illustrates how environment shaping can emerge from natural ``scarcity'' dynamics, and how dynamism can make a task seemingly more difficult, but in fact easier to learn.

\textbf{Scavenging:} The task is to find food while avoiding predators that pursue the agent, providing negative reward if caught. \textbf{Dynamism} determines the probability that the predators move to a random square, instead of moving toward the agent. \textbf{Environment shaping} determines the distance at which food appears, with food appearing closer early on. In this task, dynamism makes the task easier, allowing us to test whether training with higher dynamism still succeeds in the (harder) original task.

\textbf{Salad-making:} The task is to gather lettuce and carrots, which grow at random locations, and combine them into a salad. \textbf{Dynamism} is the frequency at which the ingredients grow at random locations. \textbf{Environment shaping} determines how close by the ingredients appear. Unlike the previous two tasks, the objects here do not move.

\textbf{Factory:} The task is to pick up wood and metal from other workers, combine them to make an axe. \textbf{Dynamism} determines the probability of workers moving to a random nearby square. \textbf{Environment shaping} controls the probability that workers move toward the agent. In this task, environment shaping has a natural interpretation as the presence of ``cooperative'' agents, while dynamism can be seen as modeling non-cooperative agents.

\textbf{Unity food collector:} This task is taken as-is from the Unity ML-Agents toolkit to evaluate on a previously proposed and standardized task. The agent must maximize the number of green spheres eaten while avoiding poisonous red spheres. \textbf{Dynamism} determines the speed at which spheres move. \textbf{Environment shaping} determines the distance at which spheres appear from the agent.

\textbf{Evaluation.} To compare agents trained under the different environment properties, we use a single consistent evaluation protocol. We use the same agent network architecture and RL algorithm for all experiments, with details in Appendix \ref{appdx:training_details}. Critically, evaluation is always done on the \textit{original} version of each task, without environment shaping or dynamism. Only the training conditions change. For each task, we generate $100$ random evaluation environments and evaluate performance of the trained policies on the evaluation environments over $10$ training seeds. 
\subsection{Difficulty of Non-Episodic RL} 
\label{sec:experiments_nonepisodic}
To demonstrate the difficulty of the non-episodic RL problem, we will compare performance for agents trained in an episodic setting against a non-episodic setting. The non-episodic version does not reset the agent to an initial state, and instead requires learning the task effectively over a single very long ``lifetime." In the non-episodic hunting task, the agent must learn to continually hunt deer over its lifetime. In the episodic version, the agent is reset to an initial state distribution after completing the task or after a fixed time horizon of $200$ time steps. Performance is measured by evaluating the trained agent on the same $100$ validation tasks, which are always episodic and static.

\begin{figure}
\vspace{-5pt}
\centering
\begin{subfigure}[b]{0.195\columnwidth}
\centering
\includegraphics[width=1.0\columnwidth]{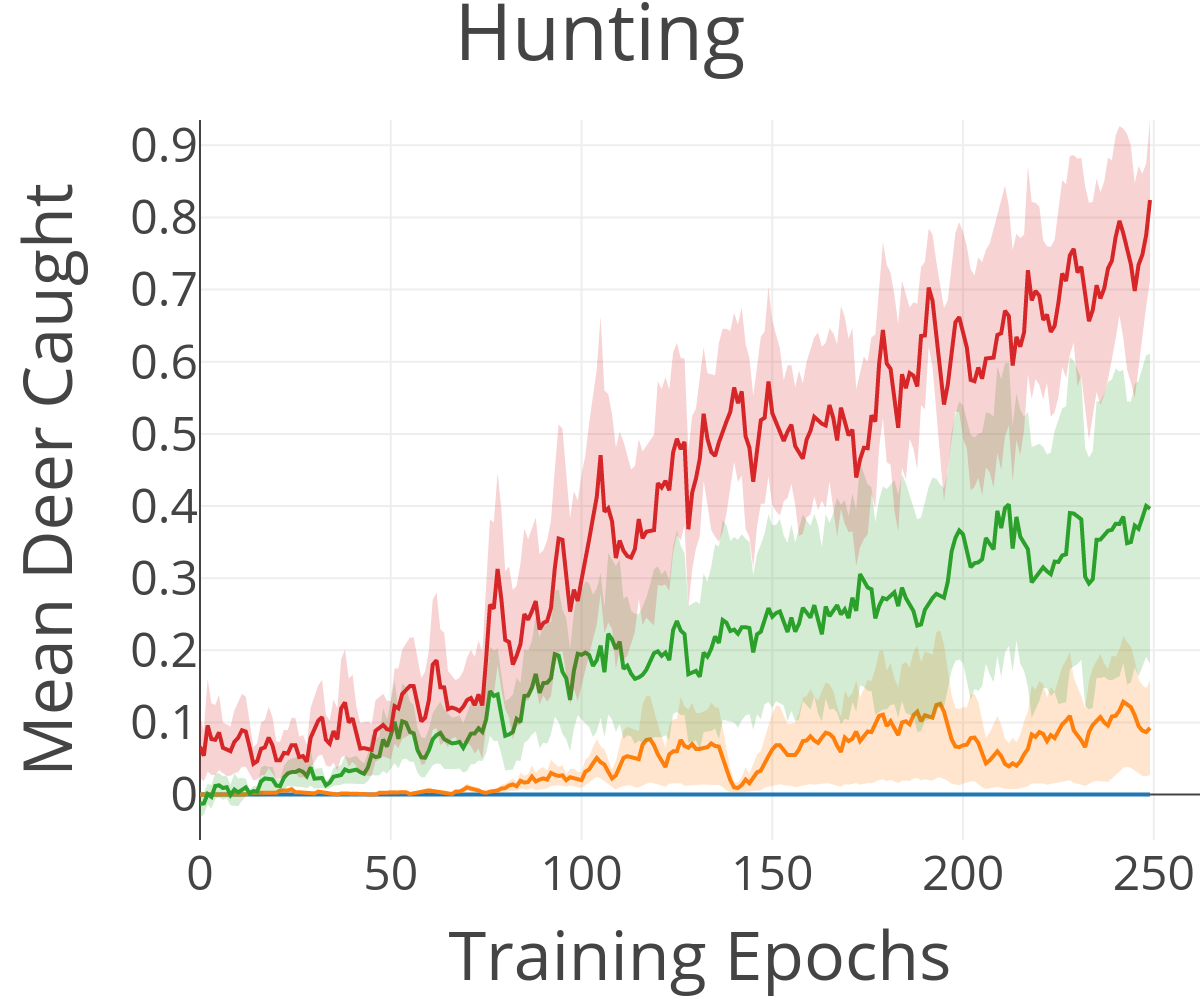}
\end{subfigure} %
\begin{subfigure}[b]{0.195\columnwidth}
\centering
\includegraphics[width=1.0\columnwidth]{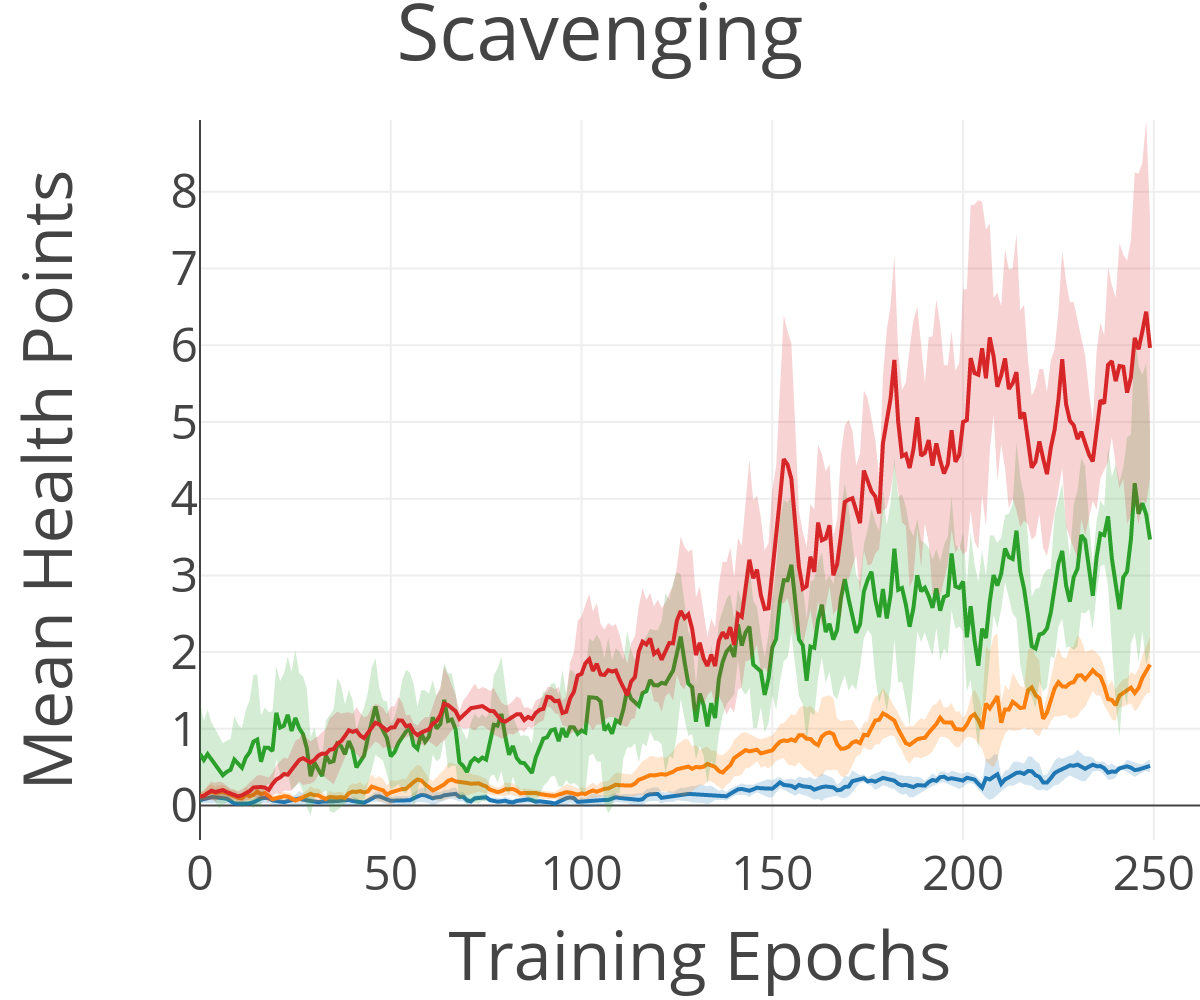}
\end{subfigure} %
\centering
\begin{subfigure}[b]{0.195\columnwidth}
\centering
\includegraphics[width=1.0\columnwidth]{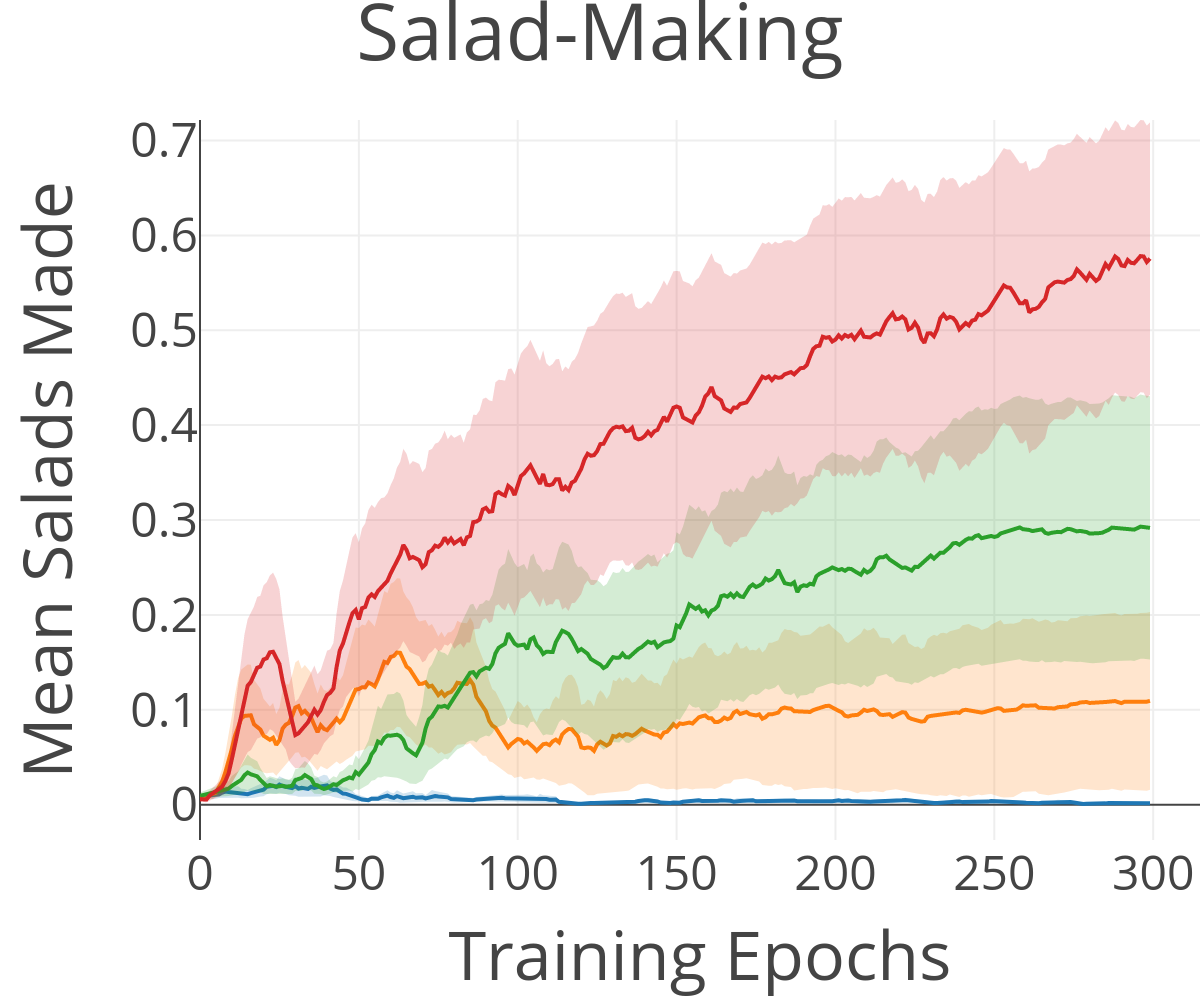}
\end{subfigure} %
\begin{subfigure}[b]{0.195\columnwidth}
\includegraphics[width=1.0\columnwidth]{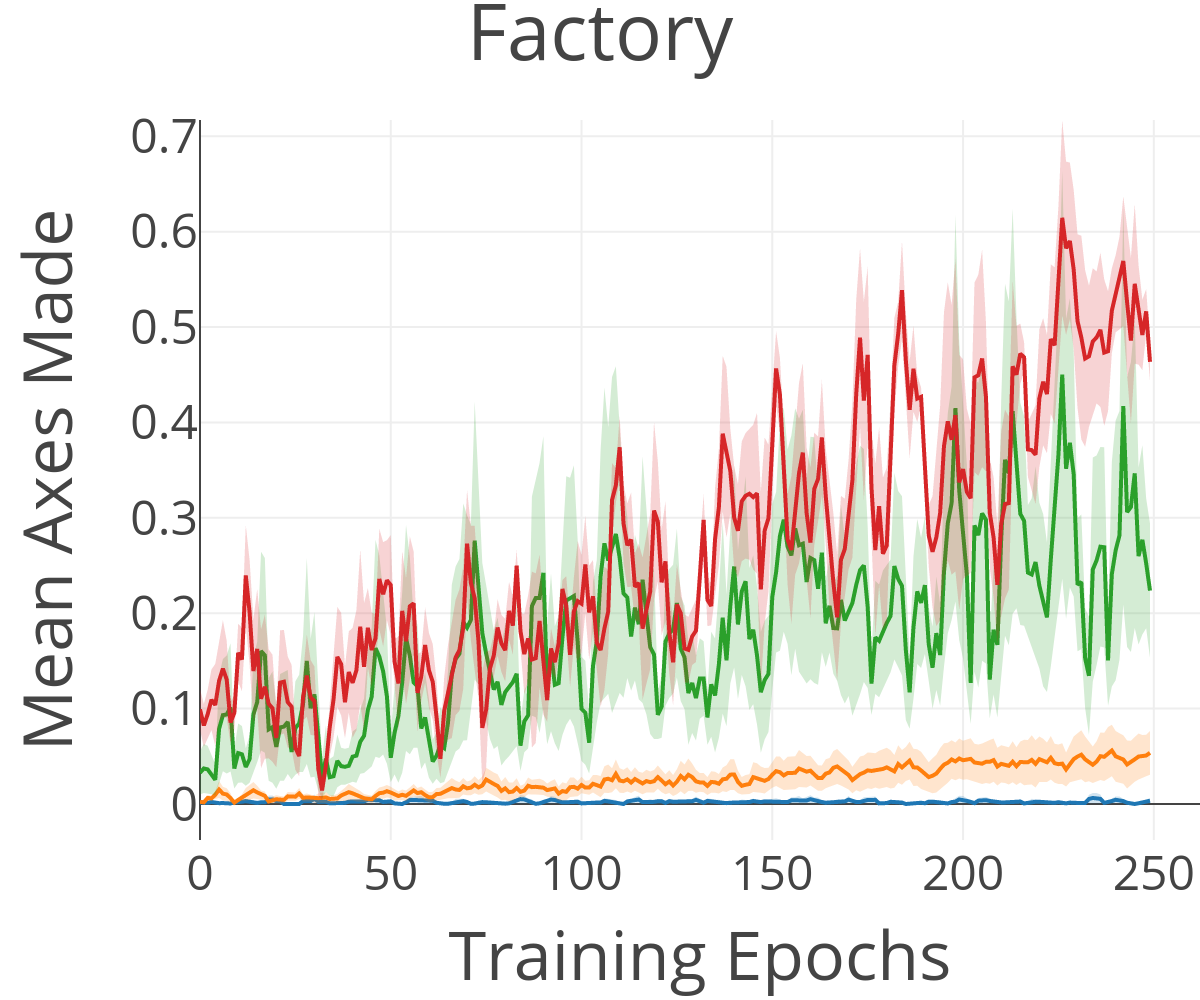}
\end{subfigure} %
\begin{subfigure}[b]{0.195\columnwidth}
\includegraphics[width=1.0\columnwidth]{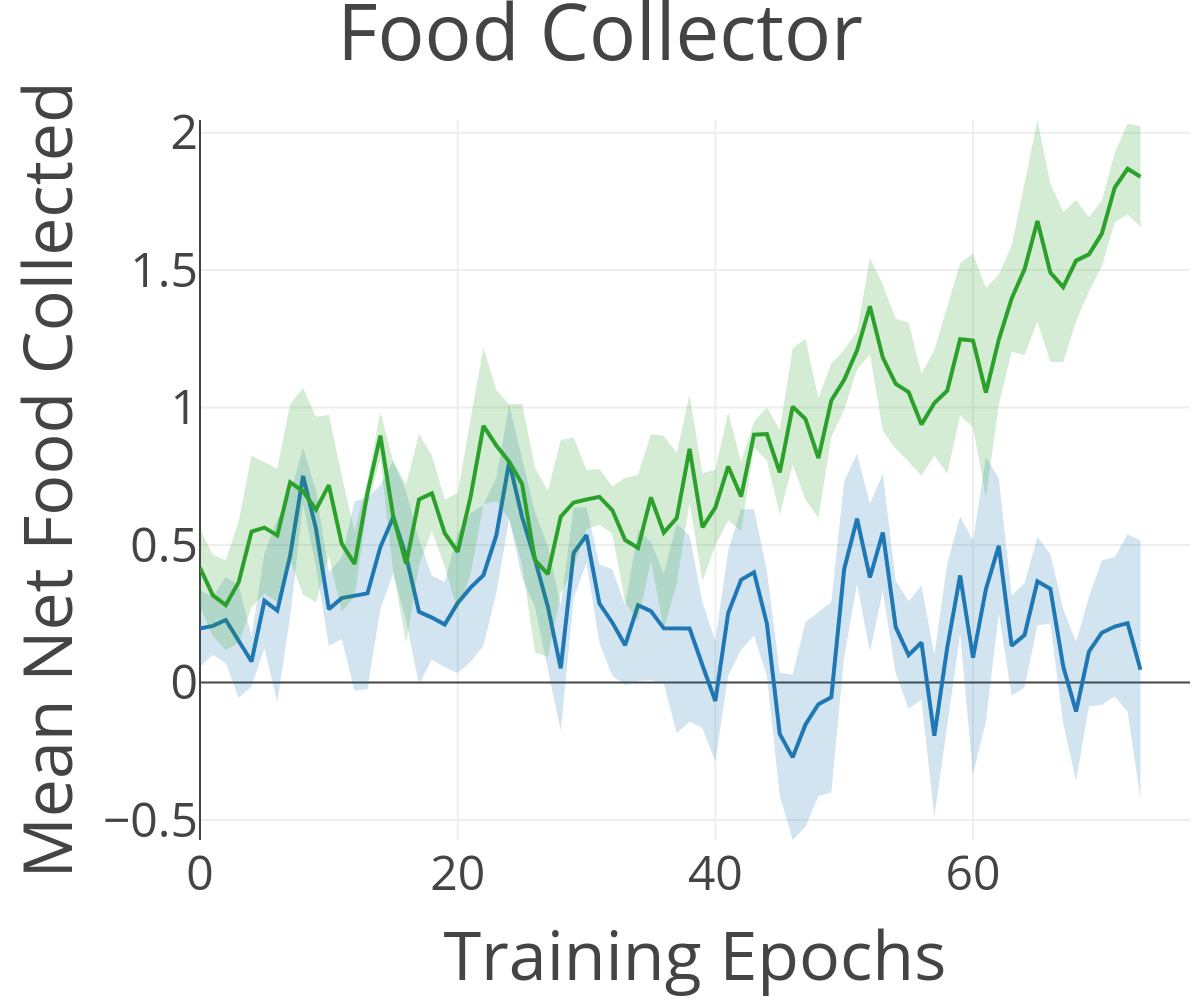}
\end{subfigure} 
\begin{subfigure}[b]{1.0\columnwidth}
\centering
\includegraphics[width=0.8\columnwidth]{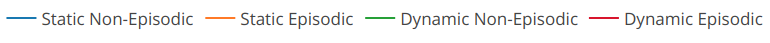}
\end{subfigure} 

    \caption{\footnotesize{Mean evaluation performance on 100 validation tasks for each training setting. Agents learning in static non-episodic environments struggle to solve the task in validation, while agents in \dynamic{} non-episodic environments are substantially more successful. Episodic learning is easier than non-episodic learning on all tasks, however non-episodic learning in \emph{dynamic} environments is more effective than episodic learning. Shaded regions represent one standard deviation over 10 seeds. Each training epoch corresponds to 5000 steps. 
    }}
    \label{fig:dynamic_static_reset_results}
    \vspace{-8pt}
\end{figure}

In Figure \ref{fig:dynamic_static_reset_results} we see, as expected, that learning tasks in static, reset-free, non-episodic environments is more difficult than a standard episodic setting. Agents trained in static non-episodic environments (blue) struggle to learn and perform worse than agents trained in static episodic environments (orange). For salad-making, the agent trained in a static non-episodic environment obtains the lowest performance ($0\%$ evaluation tasks solved), while the static episodic setting shows some signs of learning ($12\%$ evaluation tasks solved). In the static non-episodic environment, we observe that the agent frequently becomes stuck in corners of the map or in areas with no resources, as illustrated in Figure \ref{fig:nonep_vs_ep}. This can be mitigated by the properties we discuss in the next sections.

\subsection{Environment Shaping Results} \label{sec:experiments_env_shaping}
Next we study how \textit{environment shaping} can enable learning in reset-free settings with sparse reward. We will compare against no environment shaping and \textit{reward shaping}. Reward shaping is another way to alter the MDP and a common technique to enable faster learning but may be more difficult to provide in the non-episodic setting.

For all methods, the agent is given a sparse reward of $100$ each time it completes the task.  We use common techniques for reward shaping, including \emph{distance-to-goal reward shaping} \citep{Trott2019KeepingYD} and \emph{one-time reward shaping} \citep{Mataric1994RewardFF}. Distance reward shaping gives a dense distance-based reward of ($-0.01 * \text{distance to next subgoal}$) and ($1$) for resource interaction (i.e., picking up an intermediate resource). One-time reward shaping gives ($1$) reward for resource interaction and $-100$ for dropping the resource. We compare against \emph{environment shaping with sparse reward} and \emph{environment shaping with subgoal reward}, which is less shaped and only gives ($1$) for resource interaction.

\begin{figure}[tb] 
\vspace{-5pt}
\begin{subfigure}[b]{0.195\columnwidth}
\includegraphics[width=1.0\columnwidth]{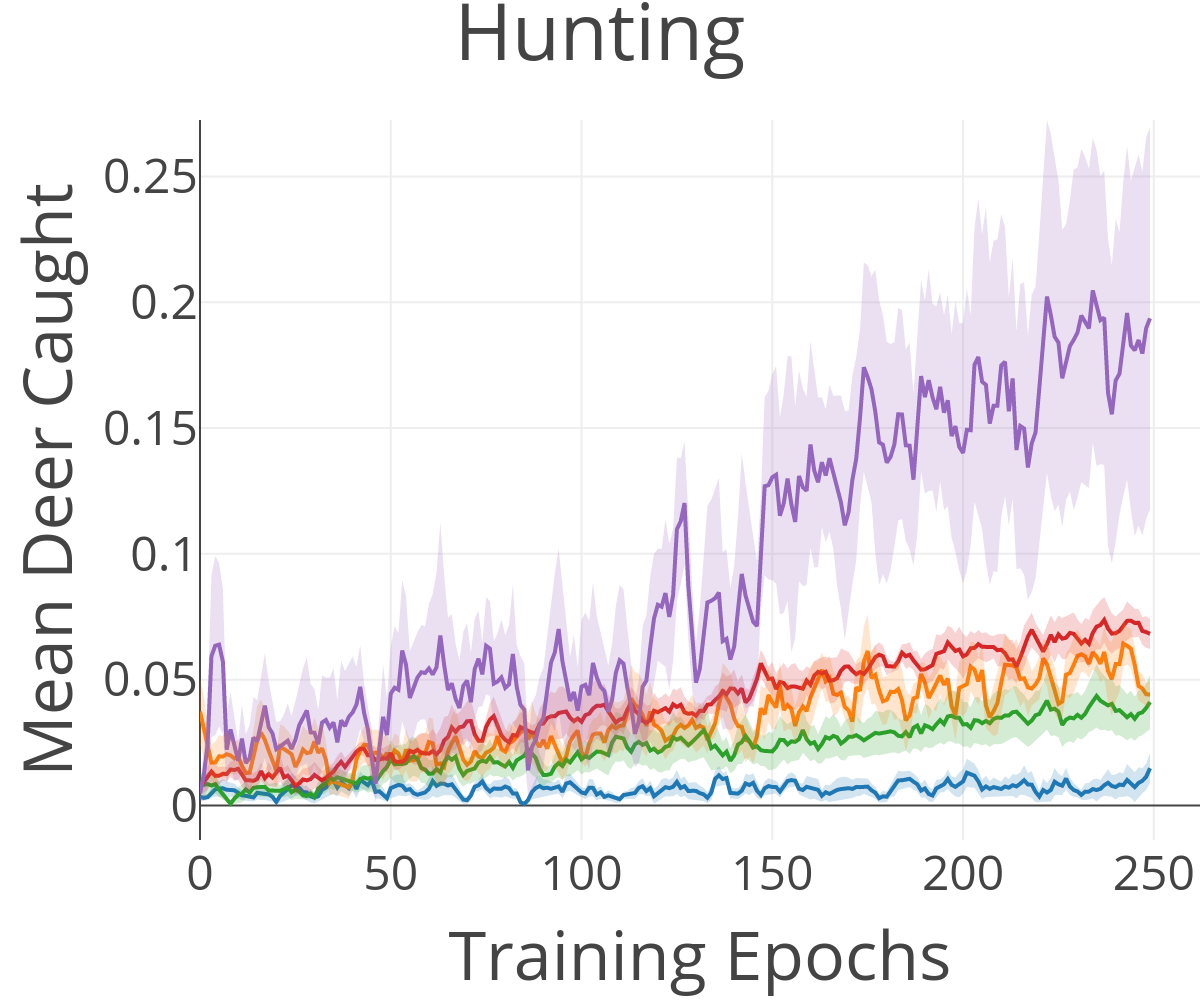}
\end{subfigure} %
\begin{subfigure}[b]{0.195\columnwidth}
\includegraphics[width=1.0\columnwidth]{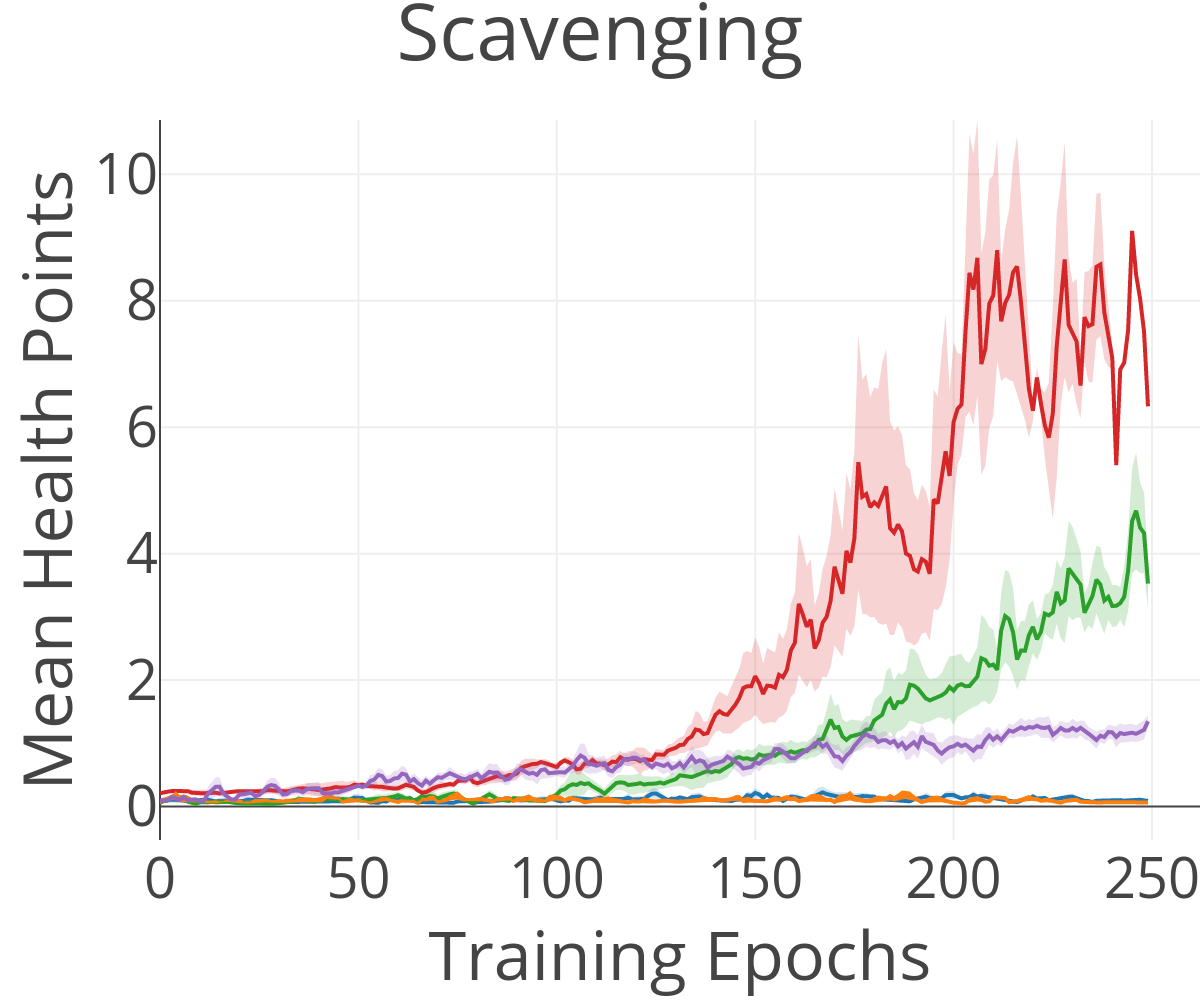}
\end{subfigure} %
\begin{subfigure}[b]{0.195\columnwidth}
\includegraphics[width=1.0\columnwidth]{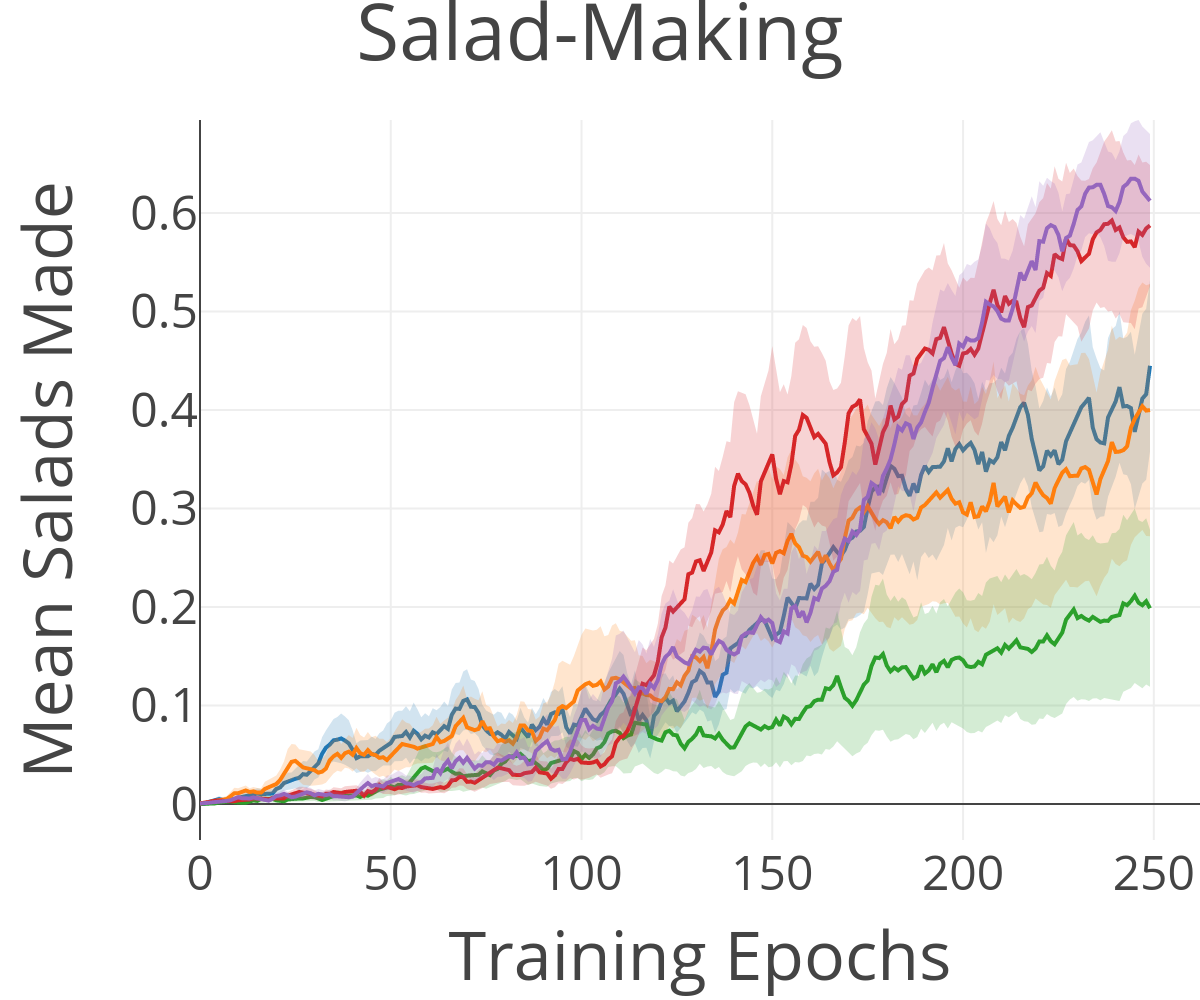}
\end{subfigure} %
\begin{subfigure}[b]{0.195\columnwidth}
\includegraphics[width=1.0\columnwidth]{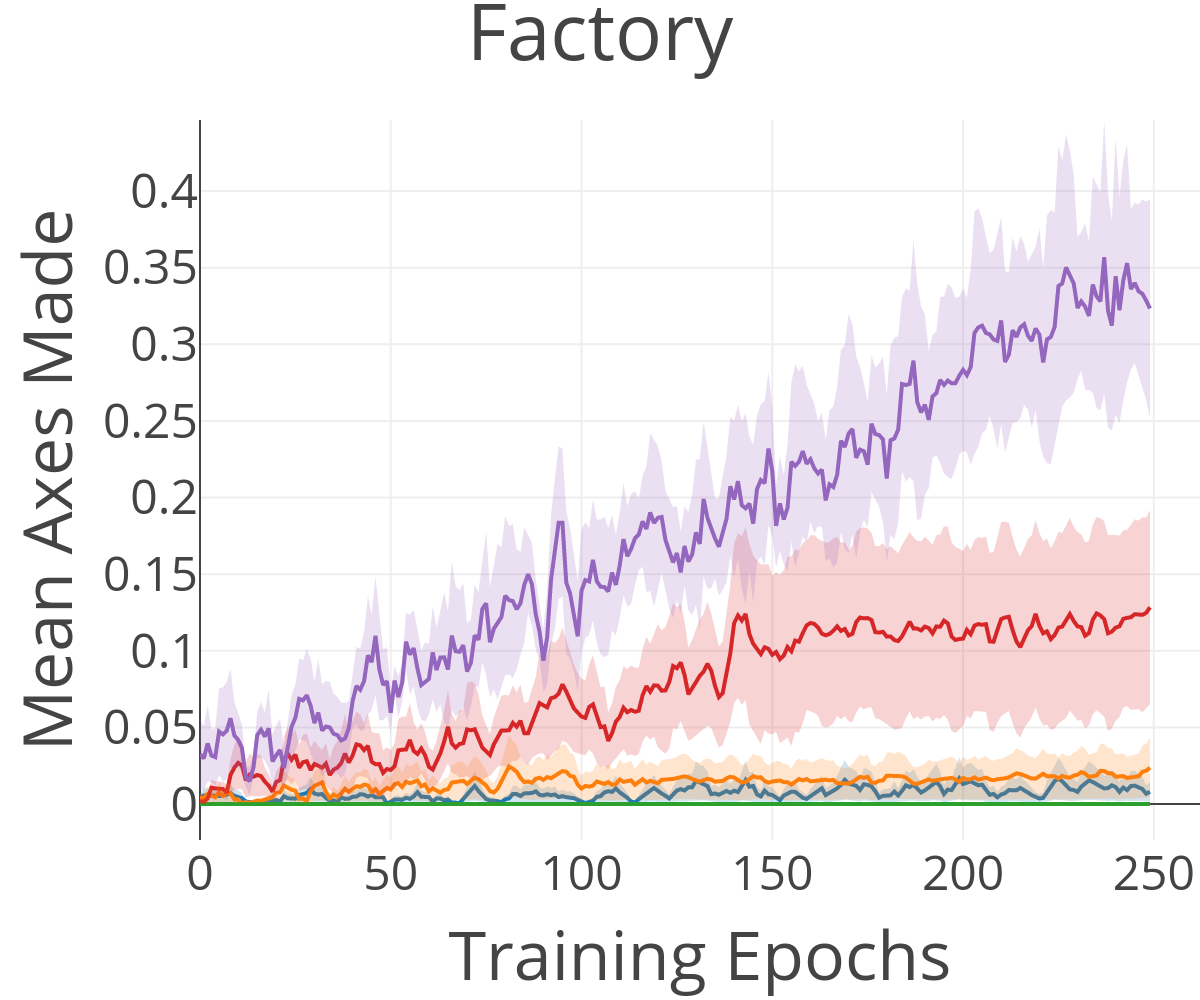}
\end{subfigure} %
\begin{subfigure}[b]{0.195\columnwidth}
\includegraphics[width=1.0\columnwidth]{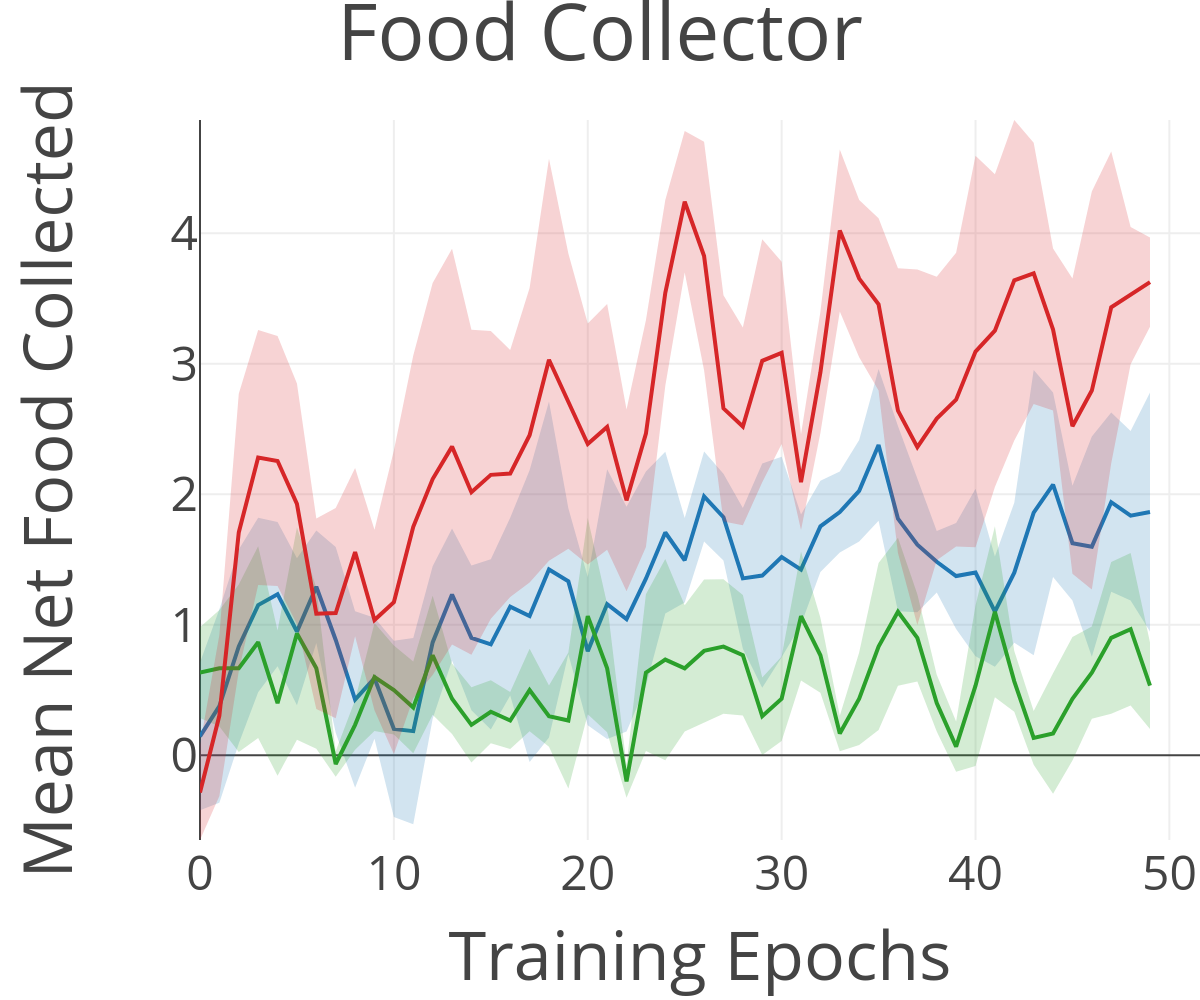}
\end{subfigure} %
\begin{subfigure}[b]{1.0\columnwidth}
\centering
\includegraphics[width=1.0\columnwidth]{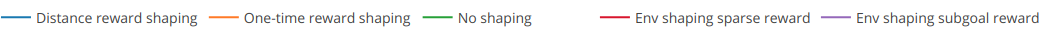}
\end{subfigure}
 \caption{\footnotesize{Evaluation performance for environment shaping in non-episodic with sparse reward and subgoal reward, in comparison to various types of reward shaping. Environment shaping more consistently facilitates effective learning. Food collector task does not have subgoals.}
 }
 \label{fig:shaping_results}
 \vspace{-10pt}
\end{figure}

We compare performance of these configurations in Figure \ref{fig:shaping_results}. We find that agents learning with environment shaping end up performing better on the test MDP across all tasks. Improper reward shaping can substantially alter the optimal policy, thereby biasing learning and resulting in a lower performance solution. These behaviors are shown illustrated in Appendix~\ref{appdx:behavior}.
As task complexity grows, so does the difficulty of constructing an unbiased shaped reward, whereas environment shaping may remain viable. We observe this difference most clearly on \textit{Hunting} and \textit{Factory}. More analysis on environment shaping robustness is in Appendix \ref{appdx:shaping_robustness}. 

\subsection{Dynamism Results}\label{sec:experiments_dynamism}
To evaluate the effect of environment dynamism on learning progress, we construct dynamic versions of our tasks by varying the probability that the environment changes at any given time, regardless of the agent's actions. This captures the fact that \emph{natural} environments change on their own, often unrelated to the agents actions. In the grid domain, we define a continuous spectrum of dynamic effects, in terms of a \emph{dynamic event probability} $p$. On the hunting task, $p$ is the probability that a deer will move to a random adjacent square on each time step. We sweep over values from [0, 1]. For the original static versions of these environments $p=0$.

\begin{figure}[b]
\vspace{-5pt}
\begin{subfigure}[b]{0.32\columnwidth}
\centering
\includegraphics[width=1.0\columnwidth]{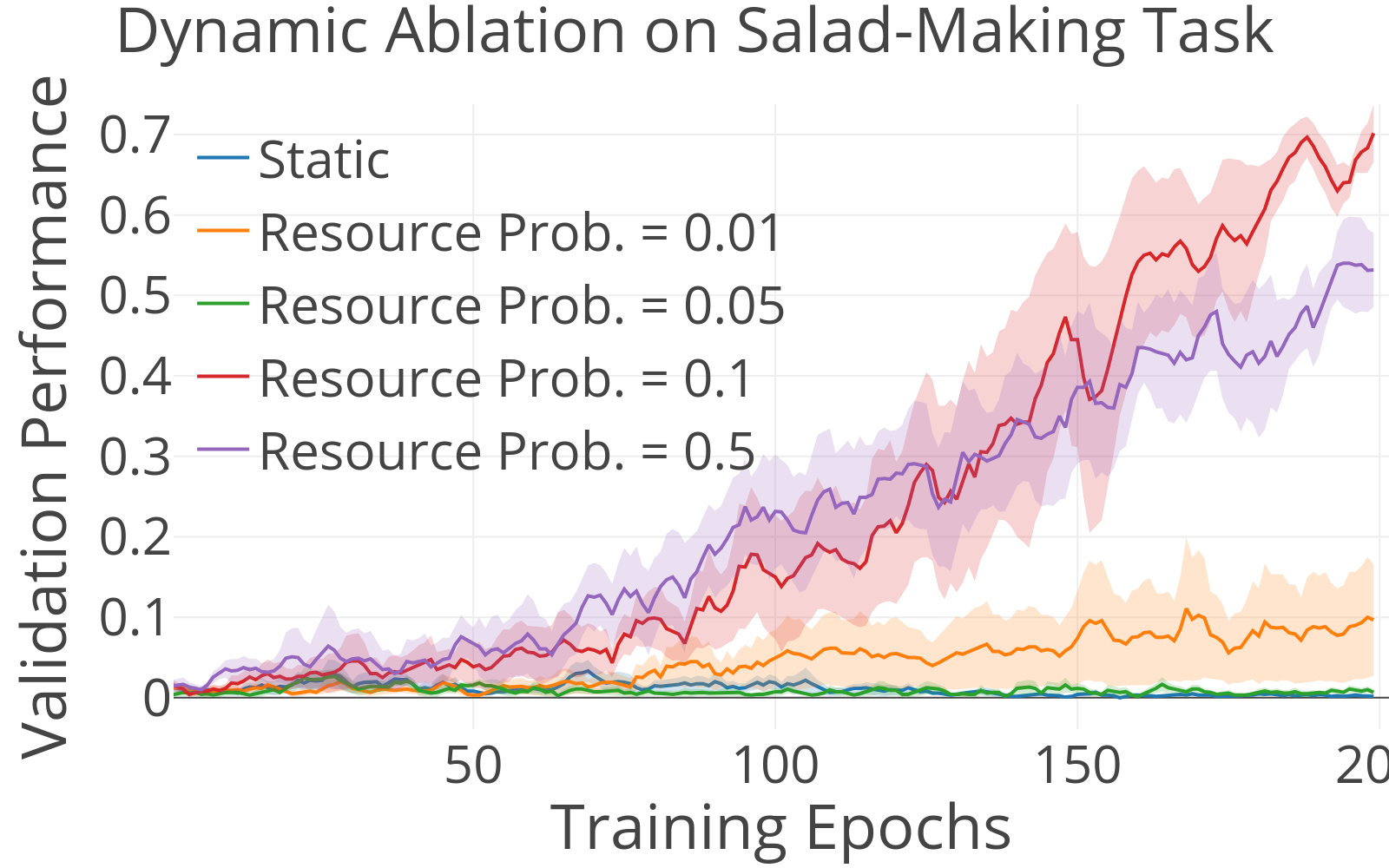}
\end{subfigure} %
\begin{subfigure}[b]{0.32\columnwidth}
\centering
\includegraphics[width=1.0\columnwidth]{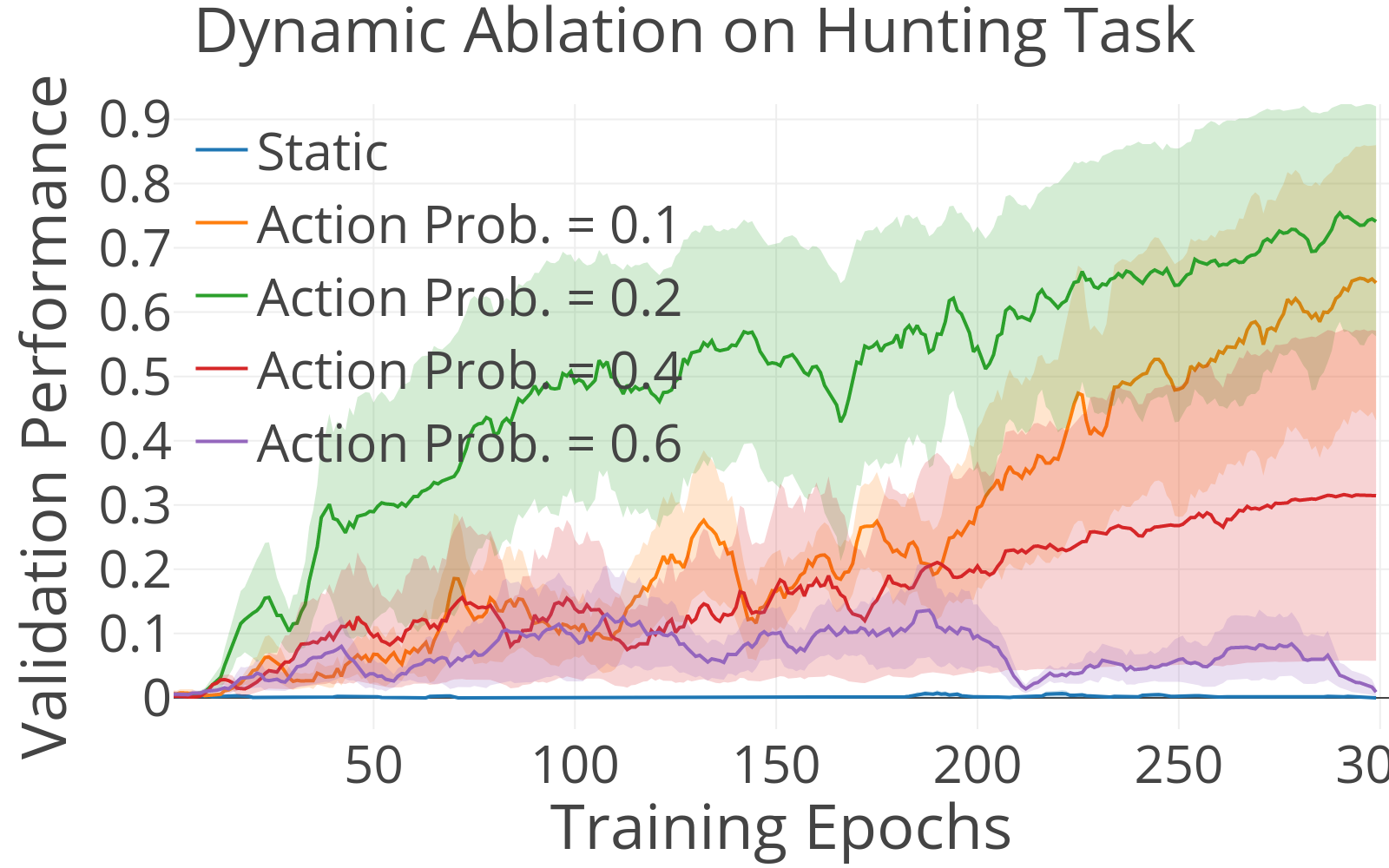}
\end{subfigure} %
\begin{subfigure}[b]{0.32\columnwidth}
\centering
\includegraphics[width=1.0\columnwidth]{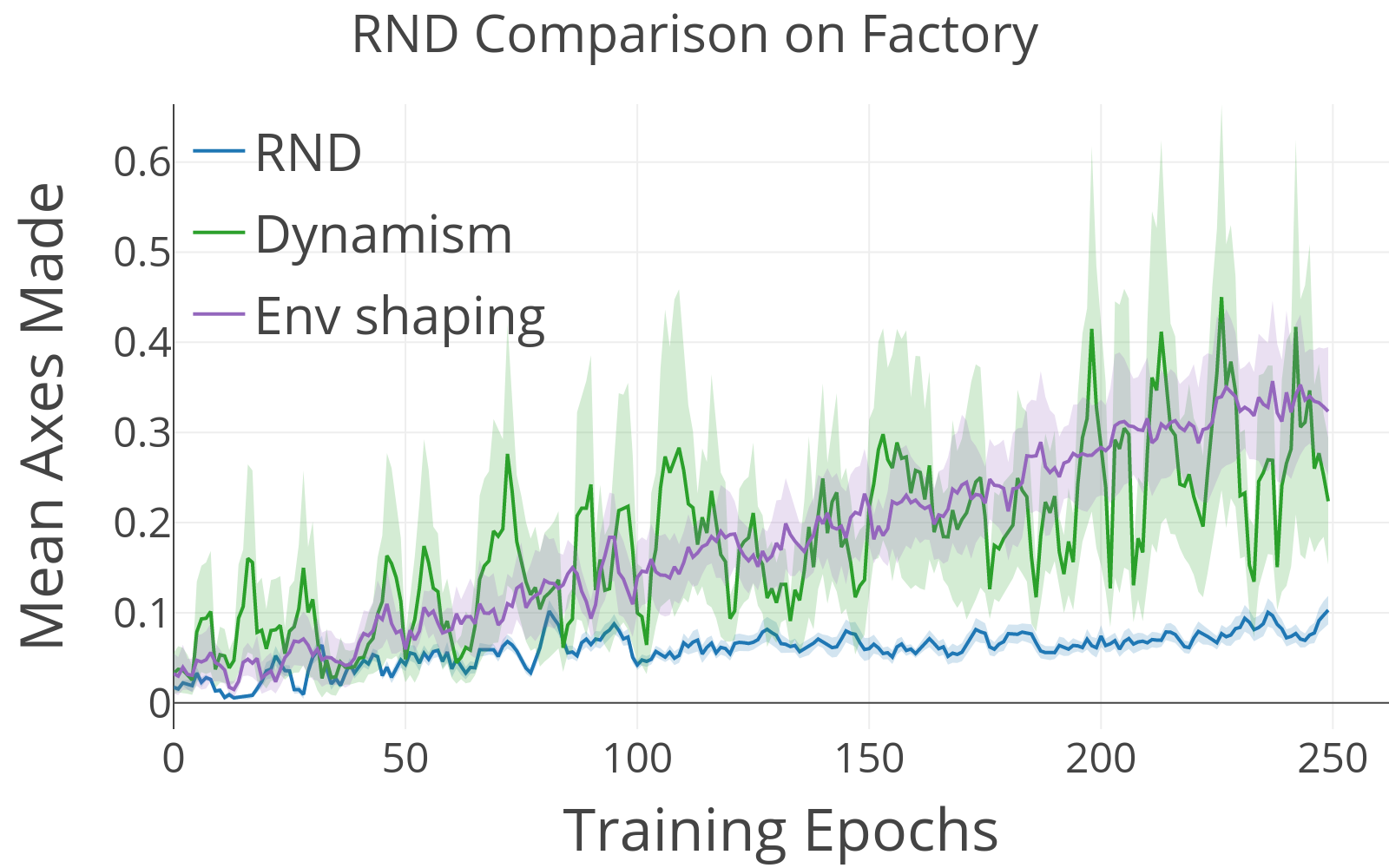}
\end{subfigure} %
    \vspace{-5pt}
    \caption{\footnotesize{Left and Midde: Evaluation performance as the \dynamic{} effect probability (i.e., resource probability and deer movement probability) is varied in the non-episodic setting. The agent is able to solve the evaluation tasks given the dynamic training environment without using resets.  For all tasks, the standard static environment does not allow for effective learning, but \dynamic{} environment variants allow the agent to learn the task successfully. Right: Comparison to exploration method RND.
    }}
     \label{fig:dynamic_static_results}
     \vspace{-14pt}
\end{figure}

Making the environment dynamic substantially improves performance, in \emph{both} the episodic and non-episodic setting, as shown in Figure~\ref{fig:dynamic_static_reset_results}. We see that the dynamic version increases performance to $70\%$, compared to the static non-episodic case ($0\%$). These results suggest that dynamic environments may to a large extent alleviate the challenges associated with non-episodic learning. Having both a dynamic environment and resets achieves the highest performance ($95\%$), indicating that both are helpful on their own. However, an environment that is too dynamic hinders performance, as we observe in Figure \ref{fig:dynamic_static_results}, where a dynamic effect probability of $0.5$ performs worse than $0.1$. This implies that environment dynamics represent a tradeoff: the environment should be stable enough for the agent to learn meaningful behavior, but dynamic enough to present interesting situations. Adding too much dynamism makes it difficult to plan for long sequences in the evaluation tasks.

This resembles the tradeoff typically encountered with exploration constants, for example in $\epsilon$-greedy exploration. To understand whether dynamism might fill a similar role to exploration algorithms, we compare non-episodic learning with dynamism or environment shaping to non-episodic learning with the addition of exploration bonuses, based on RND \cite{Burda2018ExplorationBR}. The results, shown in Figure \ref{fig:dynamic_static_reset_results}, show that both environment shaping and environment dynamism can lead to effective learning, while a policy trained with the RND exploration bonus, without dynamism or environment shaping, still struggles to solve the factory task. This indicates that, although the effect of dynamism may \emph{resemble} exploration, it provides a distinct benefit to the learning process, and exploration methods alone may not be enough to mitigate the challenges of non-episodic RL.

\section{Related Work}
\label{sec:related_work}
While reinforcement learning offers a very general formulation of the rational decision making problem, it is also known in its most general case to be intractable~\cite{Kakade2002ApproximatelyOA}. In practice, a variety of heuristics have been used to simplify RL problems, including reward shaping~\cite{Ng1999PolicyIU,wrro75121,brys2015reinforcement} and manual or automated design of curricula~\cite{NIPS2013_5187,curriculum, gravescurriculum,Randlv1998LearningTD,Yu:2018:LSL:3197517.3201397,DBLP:journals/corr/HeessTSLMWTEWER17,AlShedivat2017ContinuousAV,Sukhbaatar2017IntrinsicMA,Omidshafiei2018LearningTT,Florensa2017ReverseCG,Florensa2017AutomaticGG, Riedmiller2018LearningBP, Wang2019PairedOT}. The environment shaping approach that we study can be viewed as an instance of curriculum learning, however, curriculum learning is normally done in the episodic RL setting. Our main focus is to approach the difficulty of the RL problem from a different perspective -- instead of modifying the agent's reward function, we study how different properties of the environment can be more conducive to learning in the non-episodic RL setting where the agent is learning from a continual stream of experience. In particular, some real-world tasks might \emph{already} have these properties (e.g., more dynamic transitions), even if these properties are absent in standard benchmark tasks, or else it might be straightforward to modify the task to add these properties (e.g., ``environment shaping'' by altering initial state or dynamics to be more helpful). We hope that our experiments and analysis can be used to give insights into how both explicit and implicit curricula could be implemented more effectively to modify the training process for non-episodic RL.

While the non-episodic or ``reset-free'' setting, where the agent must learn over a single continuous lifetime, is in some sense the ``default'' setting in the standard RL formulation~\citep{Sutton1988ReinforcementLA}, it is less often studied in current benchmark tasks~\citep{Bellemare2013TheAL, Brockman2016OpenAIG}. Work on non-episodic learning is also related to the average-reward RL formulation~\cite{Wang2017PrimalDualL}. Most works observe that non-episodic learning is harder than episodic learning~\citep{Bartlett2000EstimationAA}. Learning in non-episodic settings has also been studied from the perspective of continual learning~\citep{Ring1997}. These algorithms are often concerned with ``catastrophic forgetting''~\citep{Mccloskey1989CatastrophicII,french1999catastrophic}, where previously learned tasks are forgotten while learning new tasks. To diminish this the affects of this issue, algorithms use various techniques~\citep{Rusu2016ProgressiveNN, Schwarz2018ProgressC,shingenerativereplay,ewc, Kaplanis2018ContinualRL}. Instead our algorithmic techniques focus on modifying environmental properties during training time to enable gradually learning a task.
Prior work on RL without resets has focused on safe exploration \citep{Moldovan2012SafeEI,chatzilygeroudis2018reset} or learning a policy to reset the environment~\citep{Eysenbach2017LeaveNT,Han2015LearningCM, EvenDar2005ReinforcementLI, Zhu2020TheIO}.  In our work, rather than modifying current algorithms to handle non-episodic settings more effectively (e.g., by simulating a reset), we aim to determine under which conditions non-episodic, reset-free learning is easier, and show that certain dynamic properties, which arguably are more common in realistic settings than they are in standard benchmark tasks, can actually make learning without resets substantially more tractable.

Environment shaping and dynamism can be seen as ways to mitigate the challenge of exploration during training in the sparse reward non-episodic setting. There exists many exploration methods ~\citep{Tang2016ExplorationAS, Bellemare2013TheAL, Houthooft2016VIMEVI, Pathak2017CuriosityDrivenEB, Osband2016DeepEV, Ecoffet2019GoExploreAN}, however, our work is less concerned with identifying \emph{algorithmic} features that can facilitate learning, and more with identifying \emph{environment} properties that are conducive to learning. 

\section{Discussion}
In this paper, we study how certain properties of environments affect reinforcement learning difficulty. Specifically, we examine the challenges associated with non-episodic learning without resets and simple, sparse ``fundamental drive'' reward functions, and how these challenges can be mitigated by the presence of dynamic effects (dynamism) and environment shaping -- a favorable configuration of the environment that creates a sort of curriculum. We use the term ecological reinforcement learning to refer to this type of study, which aims to analyze interactions between RL agents and the environment in which learning occurs. While our experiments show that non-episodic RL is difficult, we observe that both environment shaping and dynamism substantially alleviate these challenges. While the benefits of environment shaping may not be as surprising, more surprising is that simply increasing the randomness of the dynamics (dynamism) can make non-episodic learning tractable. Crucially, the agents trained under these conditions are evaluated in the \emph{original} unmodified MDP, where they exhibit better performance than agents trained under the same conditions as the evaluation.

The framework of ecological reinforcement learning also points to a new way to approach the design of RL agents. While reward function design is typically considered the primary modality for specifying tasks to RL agents, ecological reinforcement learning suggests that the form and structure of the environment can help to guide the emergence and specification of skills. Combined with the guidance and curricula afforded by natural environments, this suggests that studying and systematizing the interaction between RL agents and various environment properties is an important and interesting direction for future research.

\section*{Acknowledgements}
We thank Ashvin Nair, Marvin Zhang, Kelvin Xu, and anonymous reviewers for their helpful feedback and comments on earlier versions of this paper. We thank Vitchyr Pong for help with the code infrastructure. This research was supported by ARL DCIST CRA W911NF-17-2-0181, the Office of Naval Research, and the National Science Foundation under IIS-1651843, IIS-1700697, and IIS-1700696. This research was also supported in part by computational resources from Amazon and NVIDIA.
%\newpage
\bibliography{references}
\bibliographystyle{plainnat}

\clearpage
\appendix
\section{Agent Architecture and Training}
\label{appdx:training_details}
We use the same agent network architecture and RL algorithm for all our experiments with minor modification to account for the properties we vary such as reset free RL. Agents are parameterized by an MLP. The environment grid size is $8\times 8$. The partial grid observation is flattened and processed by a 2 layer MLP of size (64, 64, 32). The inventory observation is processed by a 2 layer MLP of size (16, 16, 16). These outputs are concatenated and then processed by a final MLP of size (16, action\_dim). All layers are followed by ReLU nonlinearities except the final layer which uses a softmax to output the action distribution.

\begin{figure}[h]
    \begin{subfigure}[b]{0.5\columnwidth}
    \includegraphics[width=1.0\columnwidth]{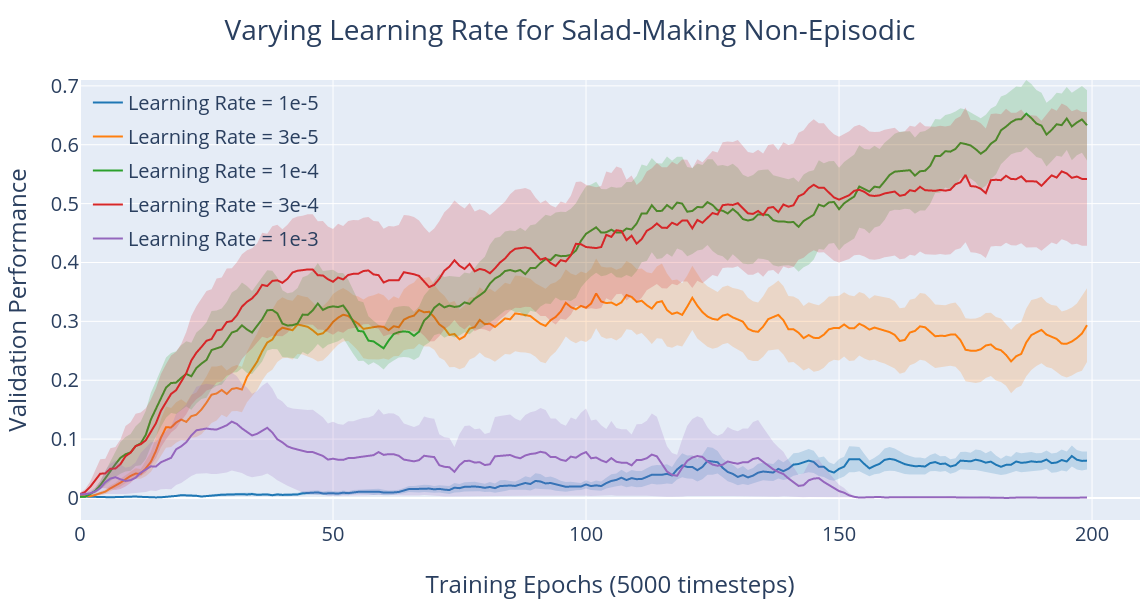}
    \end{subfigure} %
    \begin{subfigure}[b]{0.5\columnwidth}
    \includegraphics[width=1.0\columnwidth]{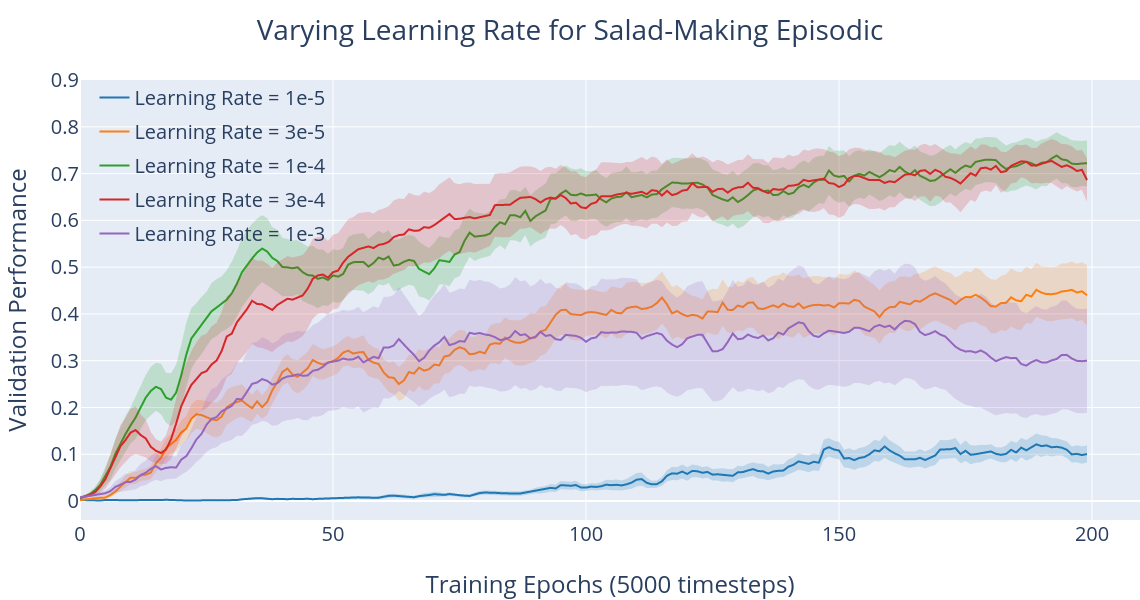}
    \end{subfigure} %
    \caption{\footnotesize{Proportion of validation tasks solved in each setting. Agents learning in static non-episodic environments struggle to learn useful behaviors, while agents learning in \dynamic{} non-episodic environments are substantially more successful. Episodic learning is easier than non-episodic learning on the first task, but non-episodic learning in \emph{dynamic} environments is almost as effective as episodic learning on the hunting task.}}
    \label{fig:learning_rate_sweep}
\end{figure}

We train the agents using double DQN \citep{Hasselt2015DeepRL} and the Adam optimizer \citep{Kingma2014AdamAM} with a learning rate selected by sweeping across a range of learning rates, with results shown in Figure \ref{fig:learning_rate_sweep}. Training is done in batch mode such that we alternate between collecting $500$ environment steps and taking $500$ gradient steps (with batch size $256$) over the replay buffer of size $5$e$5$. For episodic RL, we swept over various horizon lengths $[20, 50, 100, 200, 500]$ and found a horizon length of $200$ to work the best. We also tried setting the horizon length very short ($20$ and $50$) to help with the episodic methods but found no effect. We use epsilon greedy exploration for the policy where epsilon starts at 1 and decays linearly by $0.0001$ each timestep to $0.1$. For each training method we run $10$ random seeds. A single experiment can be run on a standard CPU machine in less than one hour.

\section{Environment Details}\label{appdx:env_details}
The grid-like environment for the first four tasks is an $N\times N$ grid that is partially observed. The agent receives a local egocentric view around it, represented by the shaded region in Figure \ref{fig:environment_example}, which is a $5\times5\times C$ grid, where $C$ is the number of object types, and each grid position contains a one-hot vector representation of the object type. The agent can pick up and carry one object at a time and can combine two objects to construct new ones by dropping a carried object onto an existing object. For example the agent can combine wood with metal to construct an axe. The action space includes moving in the cardinal directions, picking up an object, and dropping an object.

The Unity Food Collector environment has a continuous state space and the agent receives a raycast partial observation of its surrounding field of view. The action space is 27-dimensional, with separate action streams for forward, lateral, and rotational movement.
\section{Environment Shaping} \label{appdx:theory_env_shaping}
\begin{table}[h]
\begin{subtable}[t]{0.45\columnwidth} 

\centering
    \scalebox{0.7}{
        \begin{tabular}[t]{c|c}
             Env. Type & First Reward  \\
              &  Hitting Time (steps) \\
             \hline
             Static non-episodic & $908 \pm 480$ \\
             Dynamic non-episodic & $657 \pm 427$ \\
             Env. shaping non-episodic & $407 \pm 223$     
        \end{tabular}
    }
\end{subtable} %
\begin{subtable}[t]{0.45\columnwidth}
\centering
    \scalebox{0.7}{

        \begin{tabular}[t]{c|c}
             Dynamism  & Marginal State Entropy (nats) \\
             \hline
             Static Action Prob=0  & $6.2 \pm 1.2$ \\
             Dynamic Action Prob=0.01  & $7.6 \pm 0.4$ \\
             Dynamic Action Prob=0.05  &  $7.9 \pm 0.3$\\
             Dynamic Action Prob=0.1  & $8.4 \pm 0.3$ \\
             Dynamic Action Prob=0.5  & $8.6 \pm 0.5$ \\
        \end{tabular}
    }
\end{subtable}
\caption{Left: We measure the average hitting time for a uniform random policy to reach its first sparse reward in each of the environment types on the factory task. Environment shaping reduces the hitting time the most, but dynamism also reduces the hitting time. Right: We measure marginal state entropy as dynamism is varied for a random policy on the factory task where $p$ corresponds to the probability that workers move to a random square. More dynamic environments have higher marginal state entropy.}
\label{tab:hitting_time}
\end{table}

In this section we show how environment shaping (described in Section \ref{sec:properties_shaping}) can lead to faster learning and improves performance in an unbiased manner in the original unshaped environment. Shaping can also modify the dynamics but for this analysis we will assume shaping modifies the initial state distribution. To analyze the non-episodic case we will segment the learning process on the modified MDP $\trainmdp$ with initial state distribution $\shapedinitialstate(s)$ into a sequence of MDPs $\shapedmdps$ with initial states $\shapedinitialstates$ such that the terminal state of $\trainmdp_{j}$ becomes the initial state of  $\trainmdp_{j+1}$ and the optimal policy $\optpolicy_{j}$ trained on $\trainmdp_{j}$ becomes the starting policy on $\trainmdp_{j+1}$. Shaping means that $\shapedinitialstate_{1}$ has changed but as the agent learns and executes actions, the environment state and local dynamics change (such as consuming the easy deer in the hunting task in Figure \ref{fig:environment_example}). This can be modelled by a transition to a new MDP with new initial state. In this way we are segmented the learning process into a sequence of MDPs. We would like to know how fast it will take to learn a good policy in each of $\shapedmdps$ and how well the last trained policy $\optpolicy_{k}$ will perform in $\testmdp$.

We build off results from \citet{Kakade2002ApproximatelyOA,Agarwal2019OptimalityAA} which tell us how well a policy trained from $\shapedinitialstate(s)$ will perform in the MDP with initial state distribution $\initialstate(s)$. \citet{Agarwal2019OptimalityAA} gives an iteration complexity for projected gradient descent on the expected discounted return, $V^{\policy}(\initialstate)$, for a policy trained on initial state distribution $\shapedinitialstate$ (modified from Theorem 4.2 of \cite{Agarwal2019OptimalityAA}):
$$ \min_{t < T}\{V^*(\rho) - V^t(\rho)\} \leq \epsilon \text{ when } T > \frac{64\gamma|\states||\actions|}{(1-\gamma)^5\epsilon^2} \infnorm{\frac{d^{\optpolicy}_\rho(s)}{d^{\policy}_{\shapedinitialstate}(s)}}^2,$$
where $d^{\optpolicy}_\initialstate$ is the expected discounted state visitation density for $\optpolicy$ starting from $\initialstate$ and $d^{\policy}_{\shapedinitialstate}$ is the expected discounted state visitation density for $\pi$ starting from $\shapedinitialstate$. 

We apply this result iteratively on a sequence of MDPs $\shapedinitialstates$ (as in the case of environment shaping) and show the resulting iteration complexity can be less than running the algorithm on the original unshaped environment $\testmdp$. Between successive training steps, the optimal policy for $\trainmdp_{i}$ becomes the starting policy for $\trainmdp_{i+1}$. This transfer allows us to handle the non-episodic RL case where learning happens continually without resets by framing an environment transition from $\trainmdp_{i}$ to $\trainmdp_{i+1}$ as continuing from the previous policy.

The assumptions on the sequence of MDPs are $\forall \pi,s,a $:

\begin{equation}
     \infnorm{\frac{d^{\optpolicy}_{\shapedinitialstate_1}}{d^{\policy}_{\shapedinitialstate_1}}} \leq \delta, 
\end{equation} 
\begin{equation}
     \infnorm{\frac{d^{\optpolicy_{i+1}}_{\shapedinitialstate_{i+1}}}{d^{\optpolicy_{i}}_{\shapedinitialstate_{i}}}}\leq n,
\end{equation} 
\begin{equation}
     \shapedinitialstate_{k} = \initialstate.
\end{equation} 
Assumption (1) means that the first environment $\trainmdp_1$ is easy to learn in. The starting policy will be close to optimal in terms of having an expected visitation density close to the optimal policy which means that learning complexity to learn the optimal policy would be bounded by $\frac{64\gamma|\states||\actions|}{(1-\gamma)^5\epsilon^2} \delta^2$ using the above Theorem. In our experiments this assumption is manifested with a MDP that starts with many rewarding states near the initial state. 
For the hunting task (Section \ref{sec:exp_tasks}) the initial environment contains an over-abundance of easy to hunt deer (that move towards the agent) such that the agent can take any action and obtain food. Assumption (2) means that successive environments in the sequence are not changing significantly from each other. This is necessary for the optimal policy in $\trainmdp_{i}$ to transfer well to the next environment $\trainmdp_{i+1}$ which means it will not take many samples to learn the optimal policy in $\trainmdp_{i+1}$. This assumption would be reasonable in the hunting example, where the easy deer are gradually eaten, thus decreasing the proportion of easy deer. Ideally, the values of $\delta$ and $n$ should be significantly smaller than the original mismatch coefficient. Assumption (3) means that the final training MDP $\trainmdp_{k}$ is equivalent to $\testmdp$.
\vspace{-6pt}
\begin{theorem} \label{thm:shaping_complexity}
The total iteration complexity for successively training across $k$ environments $\shapedmdps$ such that the final policy obtained is $\epsilon$ optimal on the $k^{\text{th}}$ environment is
$\frac{64\gamma|\states||\actions|}{(1-\gamma)^5\epsilon^2}\left[ \delta^2 + kn^2 \right].$
\end{theorem}
\vspace{-6pt}
The resulting corollary which says that environment shaping can make learning easier is:
\vspace{-7pt}
\begin{corollary}
Training on the sequence of MDPs $\shapedmdps$ will have a lower iteration complexity than training on $\testmdp$ starting from the initial policy $\policy$ if
 \mbox{$\delta^2 + kn^2 \leq \infnorm{\frac{d^{\optpolicy}_\rho}{d^{\policy}_{\rho}}}^2$}.
\end{corollary}
\vspace{-7pt}
If the initial policy $\policy$ is very different from the optimal policy, which will be especially true for sparse reward tasks (e.g., if the chance of reaching a reward in a sparse reward task is exponentially low), then the mismatch coefficient can be arbitrarily high. We can instead use environment shaping to push the agent towards high reward states, by designing an environment that falls under these assumptions.

We analyze the time complexity for the hunting task for the environment with no shaping and environment with shaping. The goal in the hunting task is to sequentially complete the subtasks of picking up an axe, hunting deer, and eating food. We assume that eating the food is the most difficult to reach state $g$ and thus will be the state which maximizes the mismatch coefficient. Since the reward is sparse we assume that the starting policy $\pi$ takes all actions with equal probability and so approximately $\frac{1}{|\actions|}$ actions move the agent successfully towards the goal. If the number of states the optimal policy needs to pass through to finish a subtask is atleast H, then with k subtasks (3 in this case), $\frac{d^{\optpolicy}_\rho(g)}{d^{\policy}_{\initialstate}(g)} \approx |\actions|^{kH}$. 

We now analyze the time complexity for a particular instance of environment shaping in the hunting task. For the hunting task, there will be 3 training MDPs $\trainmdp_{1}, \trainmdp_{2}, \trainmdp_{3}$. $\trainmdp_{1}$ is shaped such that there is readily available food close to the agent that the agent can eat without completing the previous subtasks. So in $\trainmdp_{1}$, the mismatch coefficient will then be $\frac{d^{\optpolicy_{1}}_{\shapedinitialstate_{1}}(g)}{d^{\policy_{0}}_{\shapedinitialstate_{1}}(g)} \approx |\actions|^{H}$ since the agent just needs to randomly stumble upon the food. Once the food is exhausted then the MDP transitions to $\trainmdp_{2}$ where there are readily available axes close to the agent from which the agent can easily use to hunt the deer. Then $\frac{d^{\optpolicy_{2}}_{\shapedinitialstate_{2}}(g)}{d^{\optpolicy_{1}}_{\shapedinitialstate_{2}}(g)} \approx |\actions|^{H}$ since we assume that the optimal policy from $\trainmdp_{1}$ just needs to randomly stumble upon the deer and then will act optimally thereafter to eat the food. Thus training on the sequence of MDPs will have  $\sum_{i=1}^k \frac{d^{\optpolicy_{i}}_{\shapedinitialstate_{i}}(g)}{d^{\policy_{i-1}}_{\shapedinitialstate_{i}}(g)} \approx k|\actions|^{H}$. This form of environment shaping in effect reduces the exponent on the exploration complexity in comparison to the original environment which had a mismatch of $|\actions|^{kH}$. There are other forms which can reduce the base. For example if there were easy deer which moved toward the agent instead of running away then now more than $\frac{1}{|\actions|}$ action would move the agent closer to the goal.

In our experiments we can roughly approximate $d^{\policy}_{\shapedinitialstate}(g)$ with the average first reward hitting time which means how many steps on average it would take for the policy to hit state $g$. In Table \ref{tab:hitting_time}, we observe that environment shaping significantly reduces this hitting time. This means that a random training policy would have an increased chance of finding the goal.
\section{Environment Dynamism Theory} \label{appdx:dynamism_theory}
While environment shaping modifies the initial state distribution of $\trainmdp$ to make learning easier, environment dynamism alters the dynamics function to $\modifiedtransition$ such that it has higher entropy for all transitions. We will show that this property can induce a more uniform state distribution. This is desirable because as described in the previous section, the mismatch coefficient $\infnorm{\frac{d^{\optpolicy}_\rho}{d^{\policy}_{\rho}}}^2$ can be arbitrarily high if the initial policy fails to visit states the optimal policy visits.  Thus inducing a more uniform state distribution can reduce the mismatch coefficient in the worse case and help reduce the need for exploration.
We will again use results from Theorem 4.2 of \cite{Agarwal2019OptimalityAA} but express the mismatch coefficient in terms of the new dynamics function $\modifiedtransition$ and assume the initial state distribution is now the same. For the episodic case, \cite{Agarwal2019OptimalityAA} bypassed the need for exploration by using a uniform initial state distribution such that the mismatch coefficient would not be bad in the worst case. For the non-episodic case we will argue that environment dynamism which increase the entropy of the transition function reduces the hitting time towards a uniform state distribution. This will achieve a similar effect of pushing the agent towards a variety of states that it might need to visit.

Let $\transition(s'|s,a)$ be the original environment transition function, and let $\modifiedtransition(s'|s,a)$ be the modified transition function. We further assume that for all $s, a$, that $ \max_{s'} \modifiedtransition(s'|s,a) \leq \max_{s'} \transition(s'|s,a)$ and $ \min_{s'} \modifiedtransition(s'|s,a) \geq \min_{s'} \transition(s'|s,a)$.  The intuition of this assumption is that the dynamism operation takes the most likely transition and redistributes the probability among the other transitions, increasing the probability of the least likely transition. For a particular policy $\pi$, let $\mathcal{H}_{\transition,\pi}(\states' |\states=s) = - \sum_{s'} \transition(s'|s, \pi(s)) \log{\transition(s'|s,\pi(s))} $ denote the entropy of the conditional transition distribution. In vector form, the distribution over the next state starting from state distribution $\initialstate$ and for some policy $\pi$ will be $\transition\initialstate^T$. We will show the following theorem:
\begin{theorem} \label{thm:dynamism_uniform}
For an ergodic MDP, if for all policies $\pi$ and states $s$, $\mathcal{H}_{\modifiedtransition,\pi}(\states' |\states=s) \geq \mathcal{H}_{\transition,\pi}(\states' |\states=s)$  then for uniform state distribution $\mu$ and all state distributions $\initialstate$, $D_{TV}( \modifiedtransition  \initialstate^T, \mu) \leq D_{TV}( \transition  \initialstate^T, \mu)$.
\end{theorem}

We prove Theorem \ref{thm:dynamism_uniform}. For notation simplicity we omit the conditioning on actions. Assume that distribution $\initialstate(s) = 1$. We want to show the inequality under our assumptions on the modified dynamics:
\begin{equation} \label{eq:dyn_inequality}
    \max_{s'} |\modifiedtransition(s'|s) - \mu| \leq \max_{s'} |\transition(s'|s) - \mu| .
\end{equation}
We have four cases:
\begin{enumerate}
    \item $\max_{s'}  (\modifiedtransition(s'|s) - \mu) > 0$ and $ \max_{s'} (\transition(s'|s) - \mu) > 0$
    \item $\max_{s'}  (\modifiedtransition(s'|s) - \mu) < 0$ and $ \max_{s'} (\transition(s'|s) - \mu) < 0$
    \item $\max_{s'}  (\modifiedtransition(s'|s) - \mu) < 0$ and $ \max_{s'} (\transition(s'|s) - \mu) > 0$
    \item $\max_{s'}  (\modifiedtransition(s'|s) - \mu) > 0$ and $ \max_{s'} (\transition(s'|s) - \mu) < 0$
\end{enumerate}
Assuming that case 1 is true, then Equation \ref{eq:dyn_inequality} will follow from our assumption that $\forall s \in \states, a \in \actions, \max_{s'} \modifiedtransition(s'|s,a) \leq \max_{s'} \transition(s'|s,a)$. Assuming case 2 is true, then Equation \ref{eq:dyn_inequality} will follow from our assumption that $\forall s \in \states, a \in \actions, \min_{s'} \modifiedtransition(s'|s,a) \geq \min_{s'} \transition(s'|s,a)$.

Assuming that case 3 is true, then Equation \ref{eq:dyn_inequality} will follow from our entropy assumption: $\mathcal{H}_{\modifiedtransition,\pi}(\states' |\states=s) \geq \mathcal{H}_{\transition,\pi}(\states' |\states=s) $. To simplify the proof, let $p_1, p_2$ be the value of two transitional probabilities in $\transition$. We assume that $\modifiedtransition$ will only modify these two probabilities such that the new values in $\modifiedtransition$ are $p_1 + \epsilon$ and $p_2 - \epsilon$. We then want to show that 
\begin{equation} \label{eq:dyn_case3}
    max(|p_1 + \epsilon - \mu|,|p_2 - \epsilon - \mu|) \leq max(|p_1 - \mu|,|p_2 - \mu|).
\end{equation}
We assumed that case 3 is true. This means that $ max(p_1 + \epsilon - \mu,p_2 - \epsilon - \mu| < 0$ and $max(p_1 - \mu,p_2 - \mu| > 0$. We must have that $p_1 < p_2$. Otherwise the entropy of the new distribution would be less than the original distribution. Proof:
The change in entropy is equal to 
\begin{gather*}
    \mathcal{H}_{\modifiedtransition,\pi}(\states' |\states=s) - \mathcal{H}_{\transition,\pi}(\states' |\states=s) \\
    = -(p_1+\epsilon)\log(p_1+\epsilon) - (p_2 - \epsilon)\log(p_2 - \epsilon) + p_1\log(p_1) + p_2\log(p_2) \\
    = -p_1\log\left(\frac{p_1+\epsilon}{p_1}\right) - \epsilon\log(p_1 + \epsilon) - p_2\log\left(\frac{p_2-\epsilon}{p_2}\right) + \epsilon\log(p_2-\epsilon) \\
    = -p_1\log\left(1 + \frac{\epsilon}{p_1}\right) - \epsilon \left( \log{p_1} + \log\left(1 + \frac{\epsilon}{p_1}\right) \right)- p_2\log\left(1 - \frac{\epsilon}{p_2}\right) + \epsilon \left( \log{p_2} + \log\left(1 - \frac{\epsilon}{p_2}\right) \right) \\ 
    = -\epsilon - \epsilon \log{p_1} + \epsilon  + \epsilon\log{p_2} + O(\epsilon^2) \\
    = \epsilon \log\left(\frac{p_2}{p_1}\right) + O(\epsilon^2)
\end{gather*}
where the the 2nd to last line comes from the Taylor expansion of $\log(1+\epsilon) = \epsilon + O(\epsilon^2)$ for small $\epsilon$. If $p_1 > p_2$, then the change in entropy would be negative for small $\epsilon$. Thus, we must have that $p_1 < p_2$.

For Equation \ref{eq:dyn_case3}, we now have three cases:
\begin{enumerate}[label=(\alph*)]
    \item $p_1 > \mu$ and $p_2 > \mu$
    \item $p_1 < \mu$ and $p_2 > \mu$
    \item $p_1 < \mu$ and $p_2 > \mu$
\end{enumerate}
We assume that $\epsilon$ is small such that $p_1 + \epsilon < p_2 - \epsilon$. Case a) is not possible because this would violate the case 3 assumption $max(p_1 + \epsilon - \mu,p_2 - \epsilon - \mu|) < 0$. Case b) is not possible because this would violate the assumption $max(p_1 - \mu,p_2 - \mu|) > 0$. For case c), it is clear that moving probability from $p_2$ to $p_1$ will decrease both distances such that $|p_1 + \epsilon - \mu| < |p_1 - \mu|$ and $|p_2 - \epsilon - \mu| < |p_2 - \mu|$. This proves Equation \ref{eq:dyn_case3}. For case 4, we can make a similar argument. This proves Equation \ref{eq:dyn_inequality} for all 4 cases.

The implication of this theorem is that a higher entropy transition function should bring the state distribution closer to the uniform state distribution. This is beneficial because it will reduce the mismatch coefficient in the worst case. In our experiments we measure the marginal state entropy for different levels of dynamism. As expected we find in Table \ref{tab:hitting_time}, that more dynamic environments have higher marginal state entropy. This supports our intuition that dynamism would naturally increase state coverage.

\section{Robustness of Environment Shaping vs. Reward Shaping} \label{appdx:shaping_robustness}

\begin{figure}[h]
    \begin{subfigure}[b]{0.49\columnwidth} \label{fig:wall_env_reset}
    \includegraphics[width=1.0\columnwidth]{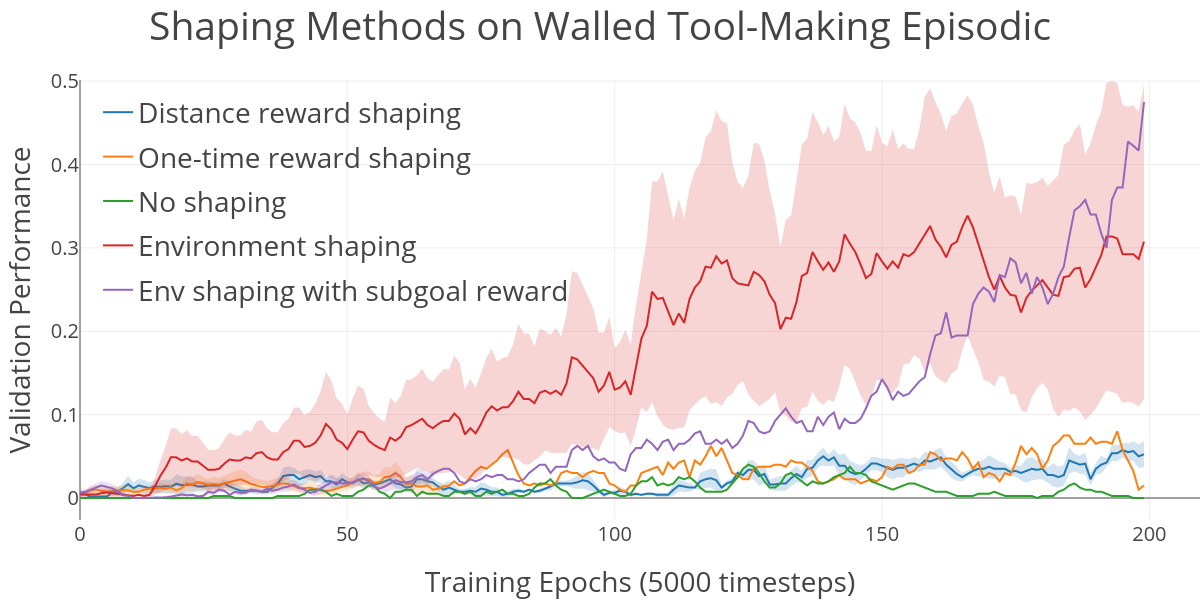}
    \end{subfigure} %
    \begin{subfigure}[b]{0.49\columnwidth} \label{fig:wall_env_resetfree}
    \includegraphics[width=1.0\columnwidth]{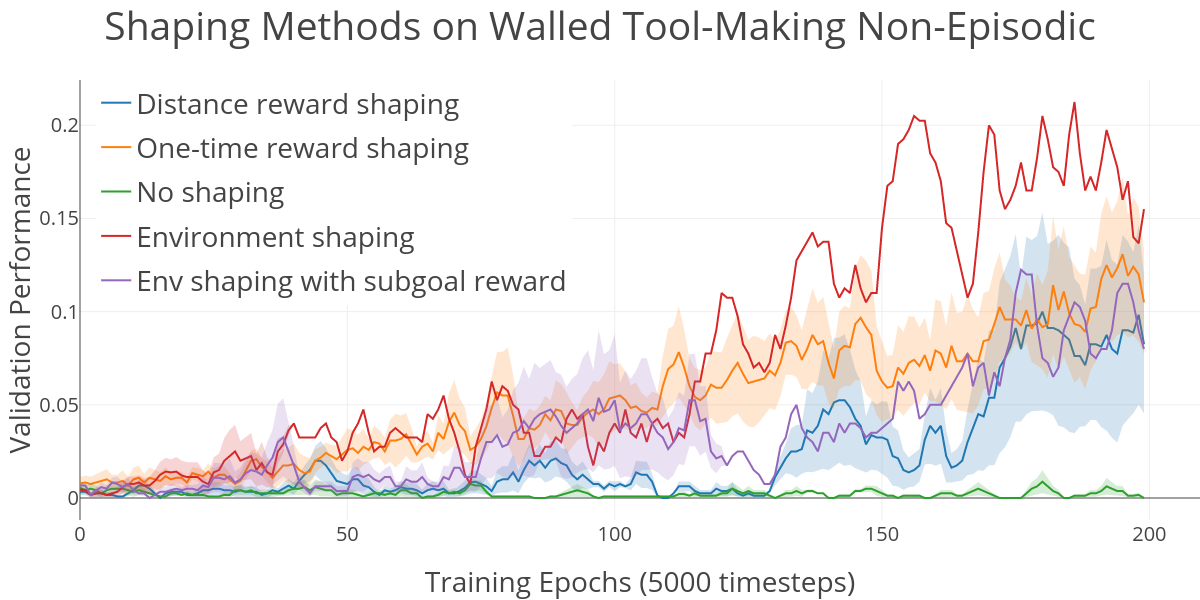}
    \end{subfigure} %
    \caption{\footnotesize{Performance of environment shaping and reward shaping on the axe-making task in an environment with wall obstacles. The distance-based reward suffers while environment shaping, despite operating off of a sparse reward, obtains peak validation performance. We find that this advantage in robustness of environment shaping is present in both the episodic and non-episodic settings, but is enhanced in the former.}}
    \label{fig:wall_env}
\end{figure}
\begin{figure}[h]
    \centering
    \begin{subfigure}[b]{0.49\columnwidth} 
    \includegraphics[width=1.0\columnwidth]{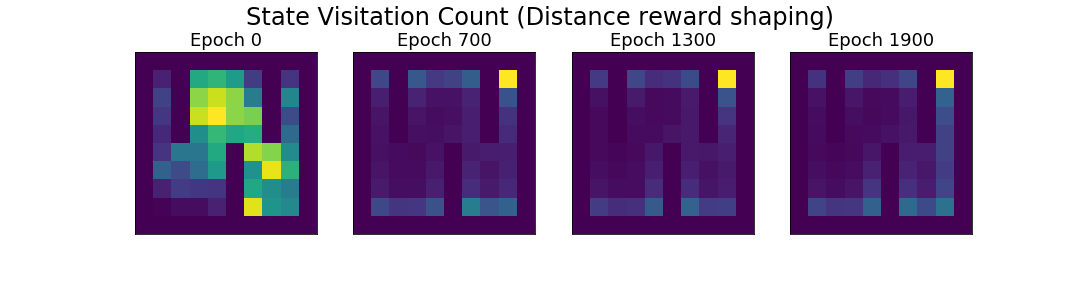}
    \end{subfigure} %
    \begin{subfigure}[b]{0.49\columnwidth} 
    \includegraphics[width=1.0\columnwidth]{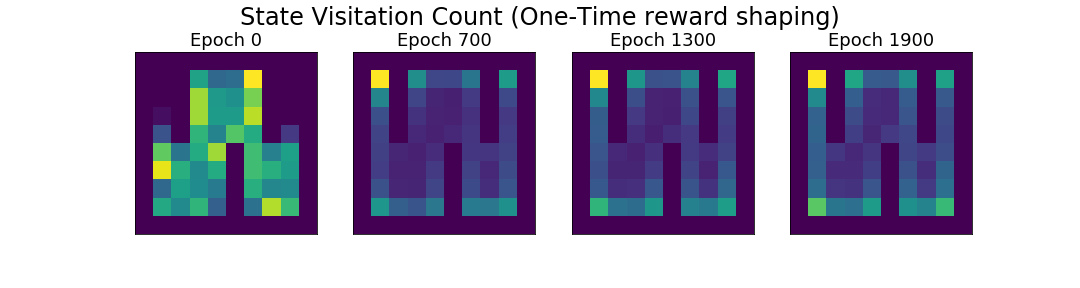}
    \end{subfigure} %
    \begin{subfigure}[b]{0.49\columnwidth} 
    \includegraphics[width=1.0\columnwidth]{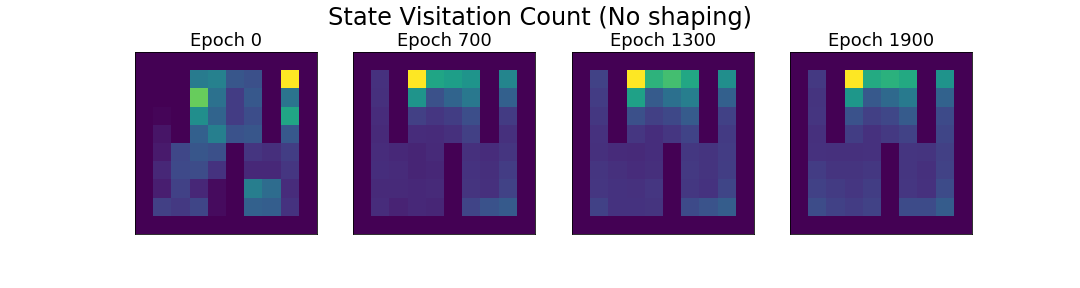}
    \end{subfigure} %
    \begin{subfigure}[b]{0.49\columnwidth}
    \includegraphics[width=1.0\columnwidth]{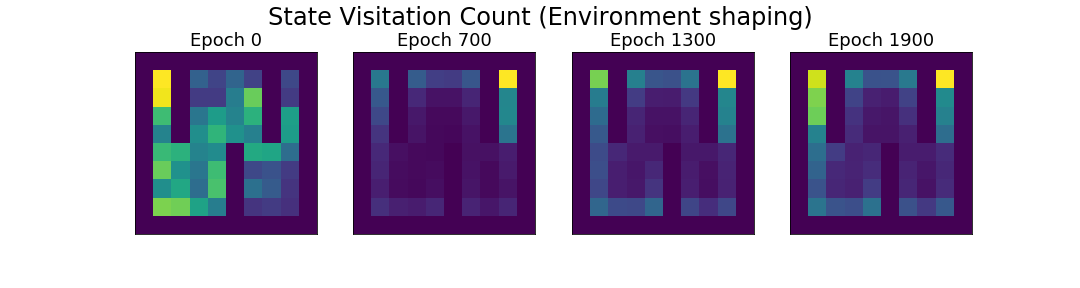}
    \end{subfigure} %
    \begin{subfigure}[b]{0.49\columnwidth} 
    \includegraphics[width=1.0\columnwidth]{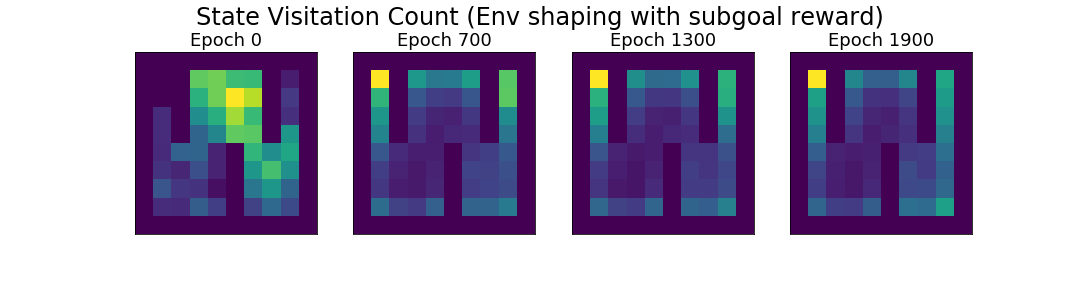}
    \end{subfigure} %
    \begin{subfigure}[b]{0.49\columnwidth} 
    \includegraphics[width=1.0\columnwidth]{imgs/exps/wall_env/heatmaps/resetfree/Distance.png}
    \end{subfigure} %
    \caption{\footnotesize{State visitation counts for the non-episodic setting visualized at 4 stages during training under the different methods of shaping. Yellow corresponds to high visitation, dark purple corresponds to low visitation. The darkest purple around the borders and in the map correspond to walls. We find that the distance-based reward shaping results in the agent getting stuck in the corners of the grid, while the one-time reward and both environment shaping methods result in the most uniform state visitation distribution over the grid during training, indicating that they were able to traverse the grid.}}
    \label{fig:wall_counts_resetfree}
\end{figure}
\begin{figure}[h]
    \centering
    \begin{subfigure}[b]{0.49\columnwidth} 
    \includegraphics[width=1.0\columnwidth]{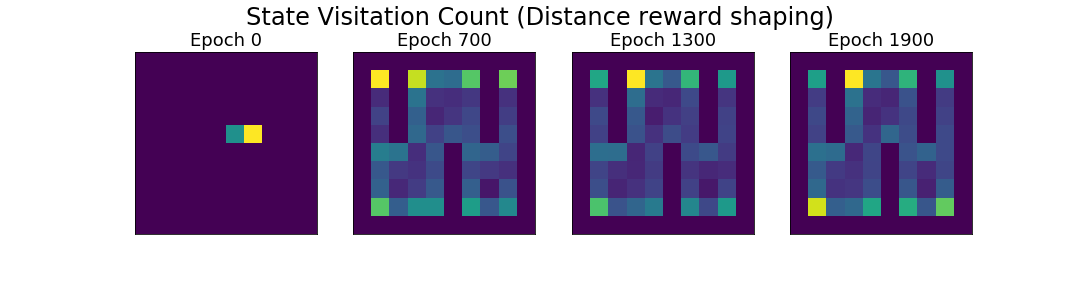}
    \end{subfigure} %
    \begin{subfigure}[b]{0.49\columnwidth}
    \includegraphics[width=1.0\columnwidth]{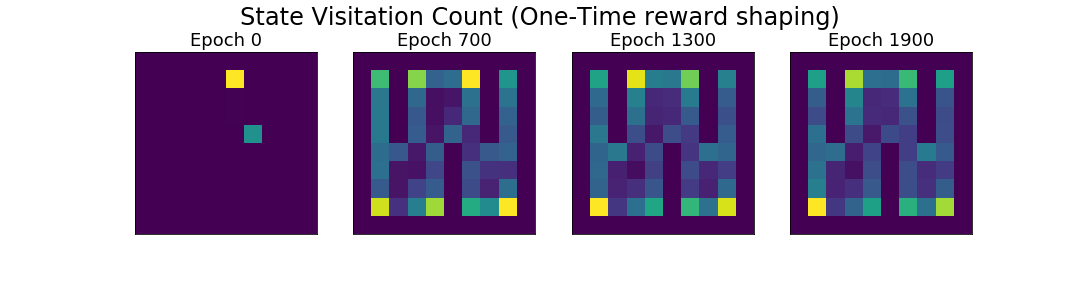}
    \end{subfigure} %
    \begin{subfigure}[b]{0.49\columnwidth} 
    \includegraphics[width=1.0\columnwidth]{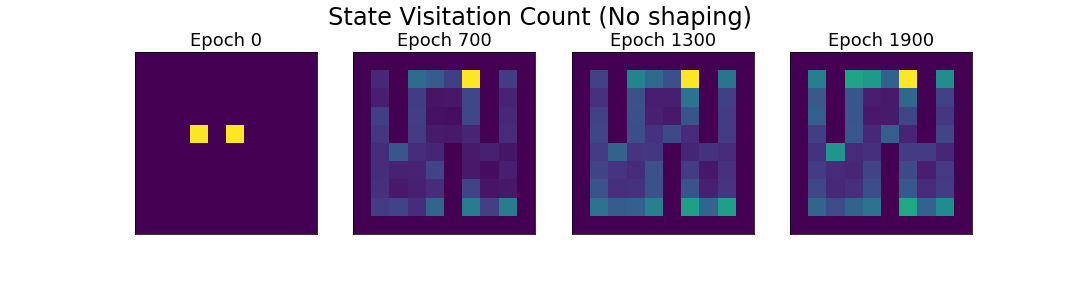}
    \end{subfigure} %
    \begin{subfigure}[b]{0.49\columnwidth} 
    \includegraphics[width=1.0\columnwidth]{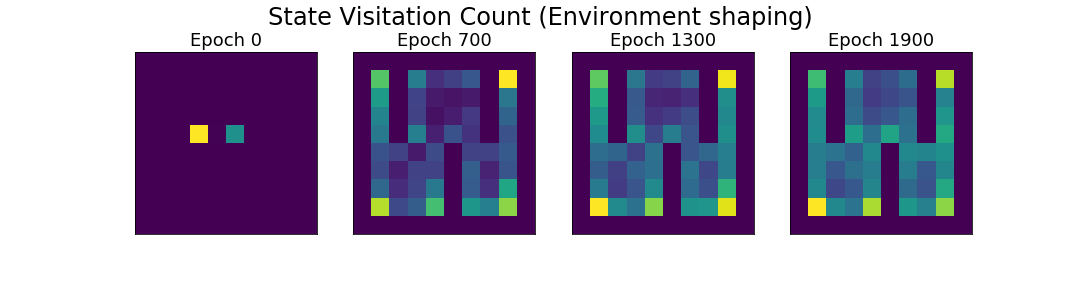}
    \end{subfigure} %
    \begin{subfigure}[b]{0.49\columnwidth} 
    \includegraphics[width=1.0\columnwidth]{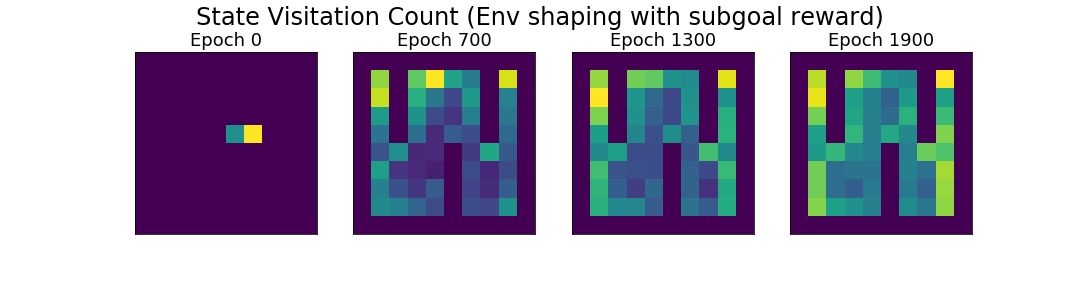}
    \end{subfigure} %

    \caption{\footnotesize{State visitation counts for the episodic setting visualized at 4 stages during training under the different methods of shaping. Yellow corresponds to high visitation, dark purple corresponds to low visitation. The darkest purple around the borders and in the map correspond to walls, which cannot be traversed by the agent. All shaping methods result in a more uniform state visitation distribution than in the non-episodic setting in Figure \ref{fig:wall_counts_resetfree}, which aligns with intuition since the resets in the episodic setting help the agent get ``unstuck.''}}
    \label{fig:wall_counts_reset}
\end{figure}

We examine the robustness of the different methods of shaping the learning of the agent by studying the performance of the agent in a more challenging environment which contain walls and are more maze-like. This makes the environment less easily navigable and provides more chances for the agent to become trapped in a particular region of the state space. We find in Figure \ref{fig:wall_env} that environment shaping is the best-performing method under this structural challenge under both episodic and non-episodic settings. However, the episodic setting demonstrates a larger gap between the performance of environment shaping and that of reward shaping. We visualize the state visitation counts of the agent under the different shaping methods in Figures \ref{fig:wall_counts_resetfree} (non-episodic) and \ref{fig:wall_counts_reset} (episodic) to understand the differences in performance. In the non-episodic setting, the distance-based reward shaping results in the agent getting trapped in corners and therefore spending a high proportion of time there. This demonstrates that reward shaping can be easier to exploit as it alters the true objective. In contrast environment shaping methods results in greater coverage of the grid.

\section{Learned Behavior}
\label{appdx:behavior}
\begin{figure}[h] 
    \centering
    \begin{subfigure}[b]{0.49\columnwidth} %\label{fig:axe_env_strip}
    \centering
    \includegraphics[width=0.6\columnwidth]{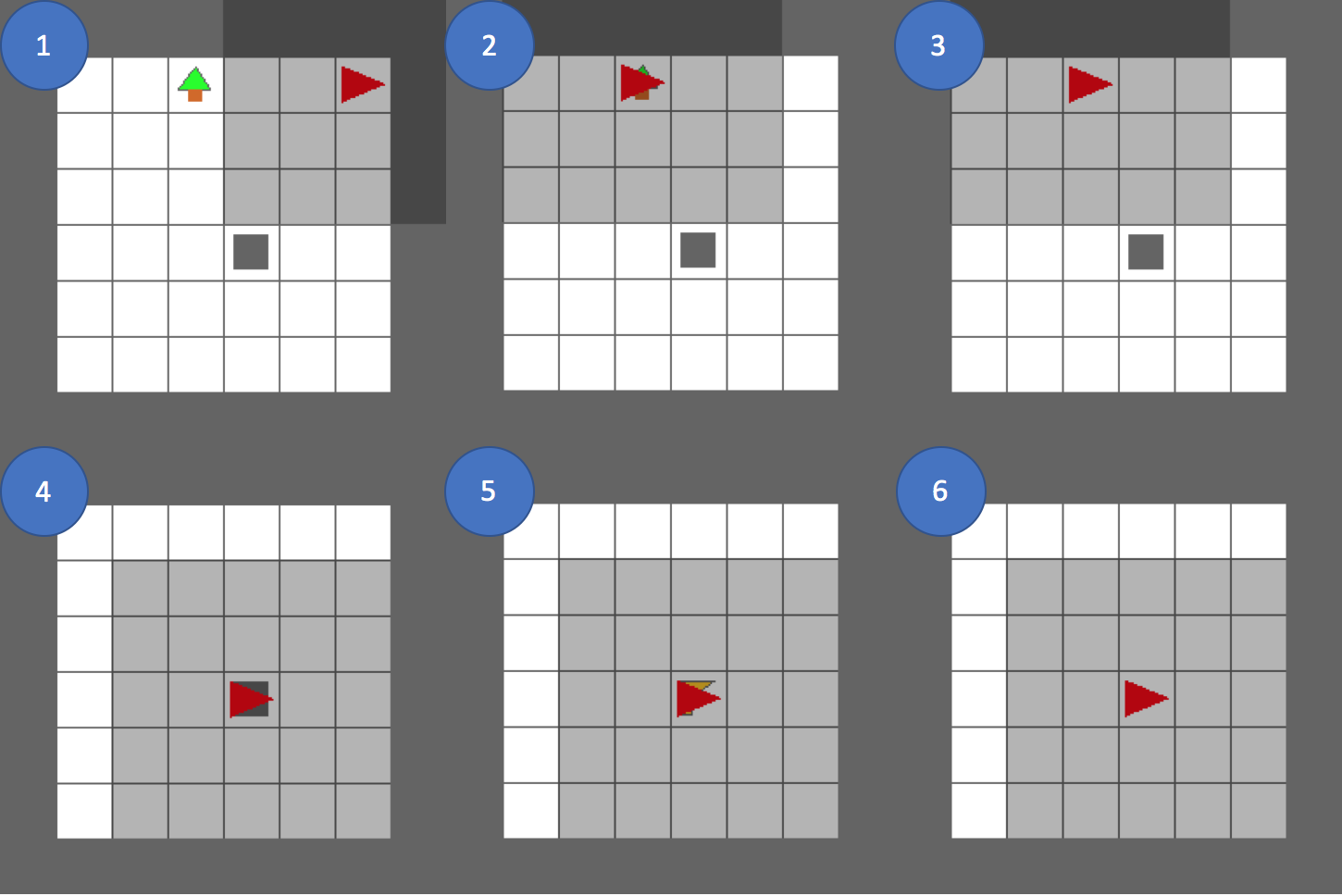}
    \end{subfigure} %
    \begin{subfigure}[b]{0.49\columnwidth} %\label{fig:axe_ot_strip}
    \centering
    \includegraphics[width=0.6\columnwidth]{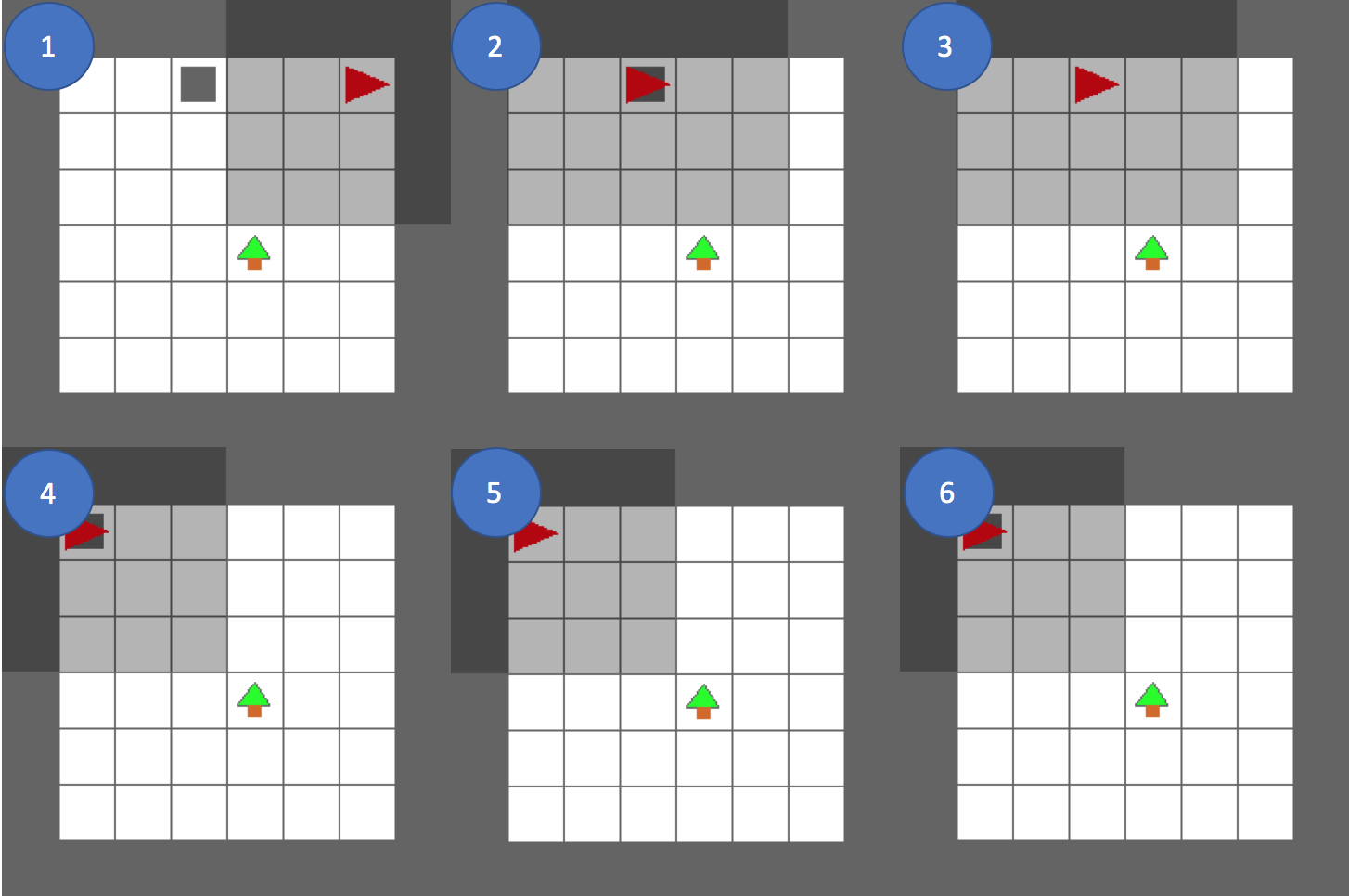}
    \end{subfigure} %
    \caption{\footnotesize{Sample trajectories on validation environments demonstrating learned behavior trained under environment shaping with sparse reward (left) as well as under shaping with the one-time reward (right), both in the non-episodic setting. The shaded region represents the agent's ego-centric partial view of the environment.}}
    \label{fig:axe_strips}
\end{figure}

\begin{figure}[t] 
    \centering
    \begin{subfigure}[b]{0.49\columnwidth} %\label{fig:deer_env_far_strip}
    \centering
    \includegraphics[width=0.6\columnwidth]{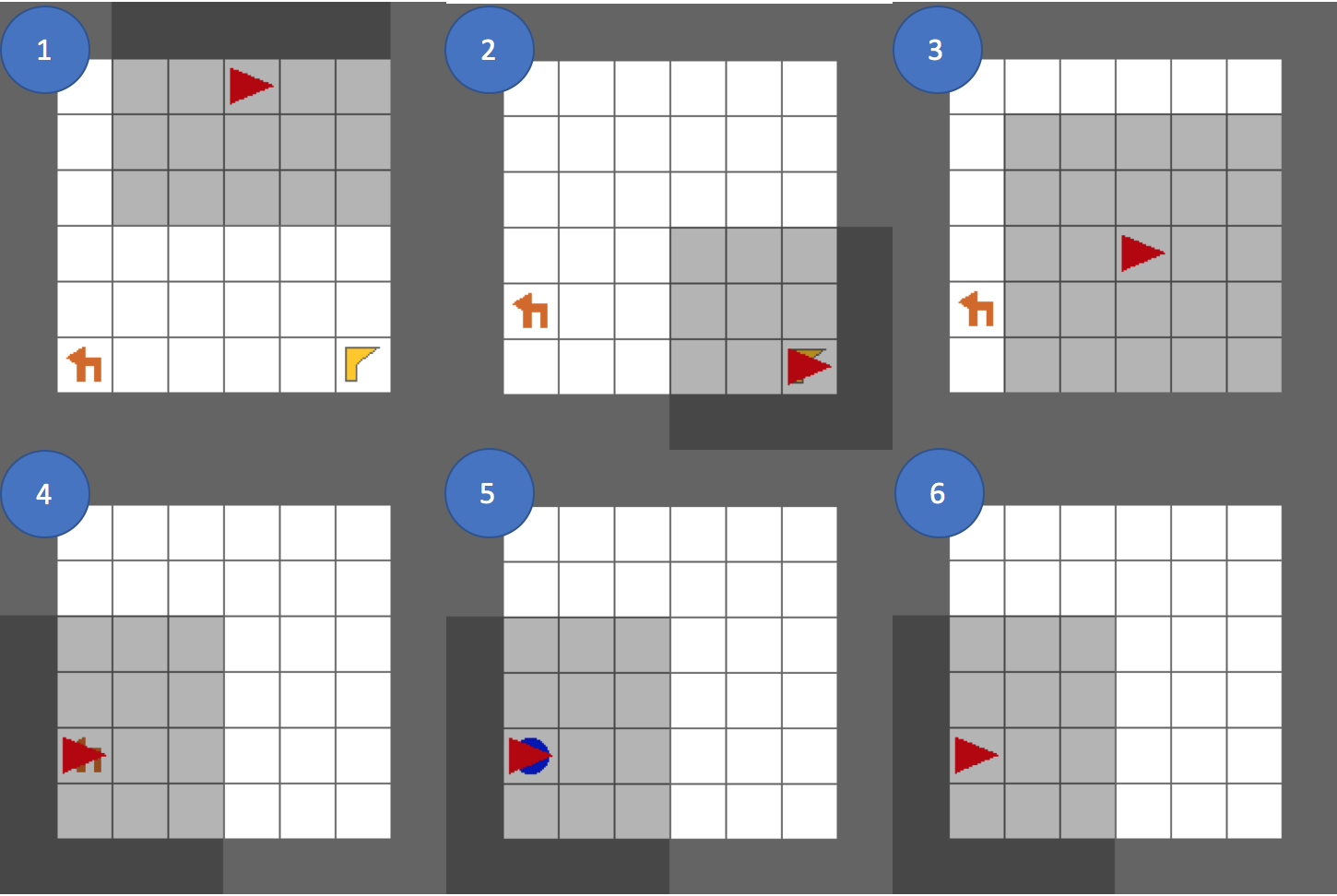}
    \end{subfigure} %
    \begin{subfigure}[b]{0.49\columnwidth} %\label{fig:deer_env_far_strip}
    \centering
    \includegraphics[width=0.6\columnwidth]{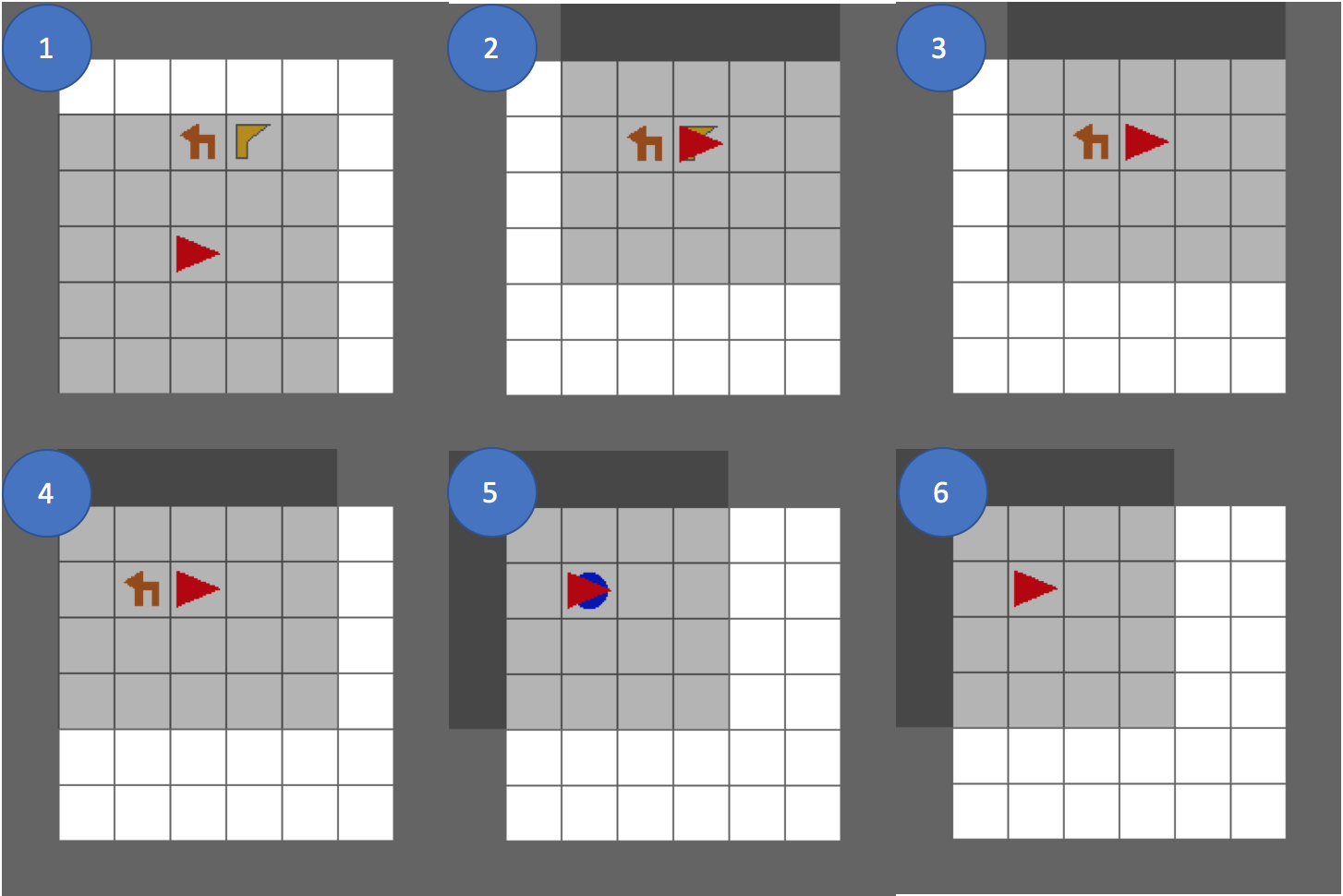}
    \end{subfigure} %
    \begin{subfigure}[b]{0.49\columnwidth} %\label{fig:deer_dist_far_strip}
    \centering
    \includegraphics[width=0.8\columnwidth]{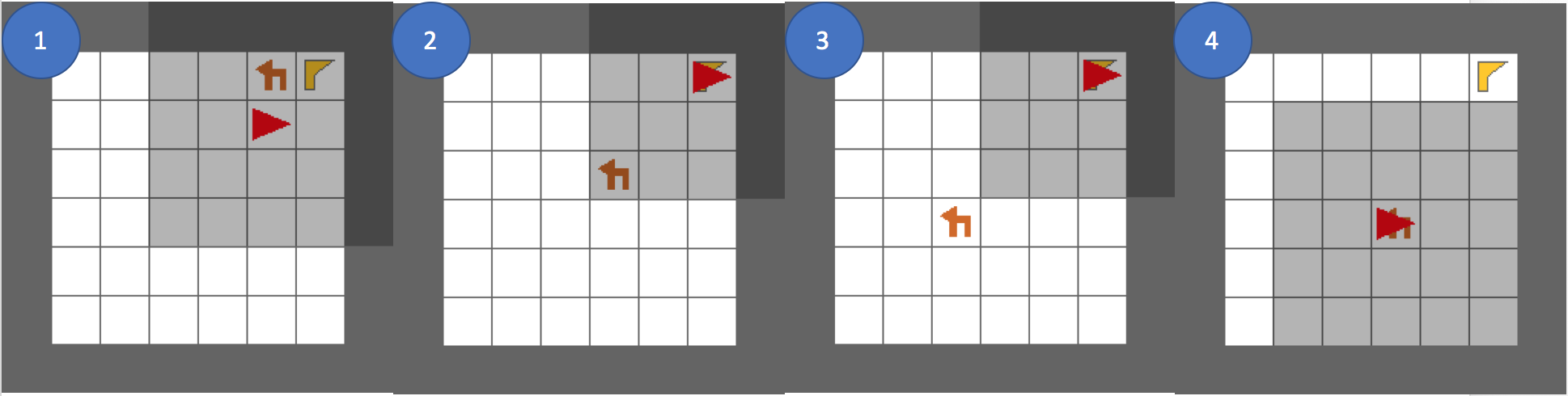}
    \end{subfigure} %
    \begin{subfigure}[b]{0.49\columnwidth} %\label{fig:deer_ot_strip}
    \centering
    \includegraphics[width=0.8\columnwidth]{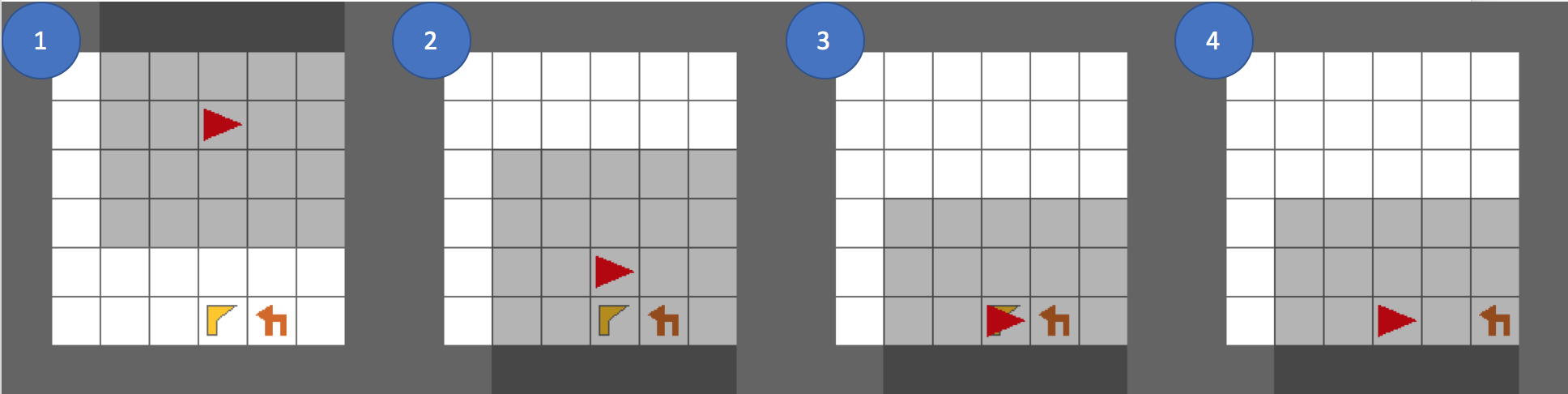}
    \end{subfigure} %
    \caption{\footnotesize{Sample trajectories on validation environments demonstrating learned behavior trained under environment shaping with sparse reward (top two), distance-based reward shaping (third), and one-time reward shaping (bottom), all in the non-episodic setting. While the environment shaped agents accomplish the desired task within 15 timesteps, biased task specification in the last two result in interpretable but suboptimal behavior. The distance-based reward shaped agent goes to the correct resources in order, but without interacting with either. The one-time reward shaped agent picks up the axe, but fails to do anything afterwards. Both reward shaped agents here fail to solve the task within the allotted 100 timesteps.}}
    \label{fig:deer_strips}
\end{figure}

We visualize the behavior of the agent under different environment training conditions. In Figure \ref{fig:axe_strips}, we demonstrate the learned behavior under environment shaping with a sparse reward and reward shaping with the one-time reward. With environment shaping, the agent accomplishes the desired task in $15$ timesteps. On the other hand, despite the fact that the one-time reward provides a reward only for the first interaction with the metal, the reward-shaped agent obtains the metal and repeatedly drops and picks it up afterwards, eventually failing to solve it within the allotted $100$ timesteps, demonstrating the biasing effect of reward shaping.
In Figure \ref{fig:deer_strips}, we visualize trajectories from the hunting environment and analyze the learned behavior of two environment-shaped agents, a distance-based reward shaped agent, and a one-time reward shaped agent. The first environment-shaped agent is able to use resources that start out on opposite sides of the world, such that they are never both in view of the agent at the same time. This is notable because the form of environment shaping used is one wherein the agent is provided with resources near it and gradually weaned off over time.

The second environment-shaped agent, while presented with a task in which the resources start out on adjacent squares, faces the challenge of the deer moving right before the agent approaches it. We observe that the trained policy is able to make a second attempt at catching the deer, and is successful. The agent trained with distance-based reward shaping displays suboptimal behavior of approaching the axe and then the deer while failing to interact with either, which can be seen as a bias resulting from a reward that incentivizes proximity to resources. Finally, the agent trained with one-time reward shaping also shows suboptimal behavior that is explained by the biases of the reward. The agent picks up the axe and then remains stationary throughout the remainder of the trajectory, failing to hunt the deer due to a reward that provides it a small reward bonus for accomplishing the first portion of the task.

\end{document}